\documentclass[sigconf]{acmart}
\usepackage{amsmath,amsfonts}
\usepackage{algorithmic}
\usepackage{graphicx}
\usepackage{textcomp}
\usepackage{xcolor}

\usepackage{xspace}
\usepackage{amsfonts}
\usepackage{bm}
\usepackage{verbatim}
\usepackage{balance}
\usepackage{graphicx}
\usepackage{multirow}
\usepackage{bbding}
\usepackage{enumitem}
\usepackage{bbm}
\usepackage{booktabs}
\usepackage[export]{adjustbox}

\usepackage{threeparttable}

\usepackage{hyperref} 

\AtBeginDocument{%
  \providecommand\BibTeX{{%
    \normalfont B\kern-0.5em{\scshape i\kern-0.25em b}\kern-0.8em\TeX}}}

\setcopyright{acmlicensed}
\copyrightyear{2025}
\acmYear{2025}
\setcopyright{cc}
\setcctype{by}
\acmConference[KDD '25]{Proceedings of the 31st ACM SIGKDD Conference on Knowledge Discovery and Data Mining V.1}{August 3--7, 2025}{Toronto, ON, Canada}
\acmBooktitle{Proceedings of the 31st ACM SIGKDD Conference on Knowledge Discovery and Data Mining V.1 (KDD '25), August 3--7, 2025, Toronto, ON, Canada}
\acmDOI{10.1145/3690624.3709277}
\acmISBN{979-8-4007-1245-6/25/08}

\usepackage{color}
\usepackage{colortbl}
\definecolor{Gray}{gray}{0.9}

\newcommand{\hide}[1]{} 

\newcommand{\vpara}[1]{\vspace{0.04in}\noindent\textbf{#1}\xspace}

\usepackage{arydshln}

\newcommand{\model}{UniGraph\xspace}

\begin{document}



\newcommand{\figleft}{{\em (Left)}}
\newcommand{\figcenter}{{\em (Center)}}
\newcommand{\figright}{{\em (Right)}}
\newcommand{\figtop}{{\em (Top)}}
\newcommand{\figbottom}{{\em (Bottom)}}
\newcommand{\captiona}{{\em (a)}}
\newcommand{\captionb}{{\em (b)}}
\newcommand{\captionc}{{\em (c)}}
\newcommand{\captiond}{{\em (d)}}

\newcommand{\newterm}[1]{{\bf #1}}

\def\figref#1{figure~\ref{#1}}
\def\Figref#1{Figure~\ref{#1}}
\def\twofigref#1#2{figures \ref{#1} and \ref{#2}}
\def\quadfigref#1#2#3#4{figures \ref{#1}, \ref{#2}, \ref{#3} and \ref{#4}}
\def\secref#1{section~\ref{#1}}
\def\Secref#1{Section~\ref{#1}}
\def\twosecrefs#1#2{sections \ref{#1} and \ref{#2}}
\def\secrefs#1#2#3{sections \ref{#1}, \ref{#2} and \ref{#3}}
\def\eqref#1{equation~\ref{#1}}
\def\Eqref#1{Equation~\ref{#1}}
\def\plaineqref#1{\ref{#1}}
\def\chapref#1{chapter~\ref{#1}}
\def\Chapref#1{Chapter~\ref{#1}}
\def\rangechapref#1#2{chapters\ref{#1}--\ref{#2}}
\def\algref#1{algorithm~\ref{#1}}
\def\Algref#1{Algorithm~\ref{#1}}
\def\twoalgref#1#2{algorithms \ref{#1} and \ref{#2}}
\def\Twoalgref#1#2{Algorithms \ref{#1} and \ref{#2}}
\def\partref#1{part~\ref{#1}}
\def\Partref#1{Part~\ref{#1}}
\def\twopartref#1#2{parts \ref{#1} and \ref{#2}}

\def\ceil#1{\lceil #1 \rceil}
\def\floor#1{\lfloor #1 \rfloor}
\def\1{\bm{1}}
\newcommand{\train}{\mathcal{D}}
\newcommand{\valid}{\mathcal{D_{\mathrm{valid}}}}
\newcommand{\test}{\mathcal{D_{\mathrm{test}}}}

\def\eps{{\epsilon}}

\def\reta{{\textnormal{$\eta$}}}
\def\ra{{\textnormal{a}}}
\def\rb{{\textnormal{b}}}
\def\rc{{\textnormal{c}}}
\def\rd{{\textnormal{d}}}
\def\re{{\textnormal{e}}}
\def\rf{{\textnormal{f}}}
\def\rg{{\textnormal{g}}}
\def\rh{{\textnormal{h}}}
\def\ri{{\textnormal{i}}}
\def\rj{{\textnormal{j}}}
\def\rk{{\textnormal{k}}}
\def\rl{{\textnormal{l}}}
\def\rn{{\textnormal{n}}}
\def\ro{{\textnormal{o}}}
\def\rp{{\textnormal{p}}}
\def\rq{{\textnormal{q}}}
\def\rr{{\textnormal{r}}}
\def\rs{{\textnormal{s}}}
\def\rt{{\textnormal{t}}}
\def\ru{{\textnormal{u}}}
\def\rv{{\textnormal{v}}}
\def\rw{{\textnormal{w}}}
\def\rx{{\textnormal{x}}}
\def\ry{{\textnormal{y}}}
\def\rz{{\textnormal{z}}}

\def\rvepsilon{{\mathbf{\epsilon}}}
\def\rvtheta{{\mathbf{\theta}}}
\def\rva{{\mathbf{a}}}
\def\rvb{{\mathbf{b}}}
\def\rvc{{\mathbf{c}}}
\def\rvd{{\mathbf{d}}}
\def\rve{{\mathbf{e}}}
\def\rvf{{\mathbf{f}}}
\def\rvg{{\mathbf{g}}}
\def\rvh{{\mathbf{h}}}
\def\rvu{{\mathbf{i}}}
\def\rvj{{\mathbf{j}}}
\def\rvk{{\mathbf{k}}}
\def\rvl{{\mathbf{l}}}
\def\rvm{{\mathbf{m}}}
\def\rvn{{\mathbf{n}}}
\def\rvo{{\mathbf{o}}}
\def\rvp{{\mathbf{p}}}
\def\rvq{{\mathbf{q}}}
\def\rvr{{\mathbf{r}}}
\def\rvs{{\mathbf{s}}}
\def\rvt{{\mathbf{t}}}
\def\rvu{{\mathbf{u}}}
\def\rvv{{\mathbf{v}}}
\def\rvw{{\mathbf{w}}}
\def\rvx{{\mathbf{x}}}
\def\rvy{{\mathbf{y}}}
\def\rvz{{\mathbf{z}}}

\def\erva{{\textnormal{a}}}
\def\ervb{{\textnormal{b}}}
\def\ervc{{\textnormal{c}}}
\def\ervd{{\textnormal{d}}}
\def\erve{{\textnormal{e}}}
\def\ervf{{\textnormal{f}}}
\def\ervg{{\textnormal{g}}}
\def\ervh{{\textnormal{h}}}
\def\ervi{{\textnormal{i}}}
\def\ervj{{\textnormal{j}}}
\def\ervk{{\textnormal{k}}}
\def\ervl{{\textnormal{l}}}
\def\ervm{{\textnormal{m}}}
\def\ervn{{\textnormal{n}}}
\def\ervo{{\textnormal{o}}}
\def\ervp{{\textnormal{p}}}
\def\ervq{{\textnormal{q}}}
\def\ervr{{\textnormal{r}}}
\def\ervs{{\textnormal{s}}}
\def\ervt{{\textnormal{t}}}
\def\ervu{{\textnormal{u}}}
\def\ervv{{\textnormal{v}}}
\def\ervw{{\textnormal{w}}}
\def\ervx{{\textnormal{x}}}
\def\ervy{{\textnormal{y}}}
\def\ervz{{\textnormal{z}}}

\def\rmA{{\mathbf{A}}}
\def\rmB{{\mathbf{B}}}
\def\rmC{{\mathbf{C}}}
\def\rmD{{\mathbf{D}}}
\def\rmE{{\mathbf{E}}}
\def\rmF{{\mathbf{F}}}
\def\rmG{{\mathbf{G}}}
\def\rmH{{\mathbf{H}}}
\def\rmI{{\mathbf{I}}}
\def\rmJ{{\mathbf{J}}}
\def\rmK{{\mathbf{K}}}
\def\rmL{{\mathbf{L}}}
\def\rmM{{\mathbf{M}}}
\def\rmN{{\mathbf{N}}}
\def\rmO{{\mathbf{O}}}
\def\rmP{{\mathbf{P}}}
\def\rmQ{{\mathbf{Q}}}
\def\rmR{{\mathbf{R}}}
\def\rmS{{\mathbf{S}}}
\def\rmT{{\mathbf{T}}}
\def\rmU{{\mathbf{U}}}
\def\rmV{{\mathbf{V}}}
\def\rmW{{\mathbf{W}}}
\def\rmX{{\mathbf{X}}}
\def\rmY{{\mathbf{Y}}}
\def\rmZ{{\mathbf{Z}}}

\def\ermA{{\textnormal{A}}}
\def\ermB{{\textnormal{B}}}
\def\ermC{{\textnormal{C}}}
\def\ermD{{\textnormal{D}}}
\def\ermE{{\textnormal{E}}}
\def\ermF{{\textnormal{F}}}
\def\ermG{{\textnormal{G}}}
\def\ermH{{\textnormal{H}}}
\def\ermI{{\textnormal{I}}}
\def\ermJ{{\textnormal{J}}}
\def\ermK{{\textnormal{K}}}
\def\ermL{{\textnormal{L}}}
\def\ermM{{\textnormal{M}}}
\def\ermN{{\textnormal{N}}}
\def\ermO{{\textnormal{O}}}
\def\ermP{{\textnormal{P}}}
\def\ermQ{{\textnormal{Q}}}
\def\ermR{{\textnormal{R}}}
\def\ermS{{\textnormal{S}}}
\def\ermT{{\textnormal{T}}}
\def\ermU{{\textnormal{U}}}
\def\ermV{{\textnormal{V}}}
\def\ermW{{\textnormal{W}}}
\def\ermX{{\textnormal{X}}}
\def\ermY{{\textnormal{Y}}}
\def\ermZ{{\textnormal{Z}}}

\def\vzero{{\bm{0}}}
\def\vone{{\bm{1}}}
\def\vmu{{\bm{\mu}}}
\def\vtheta{{\bm{\theta}}}
\def\va{{\bm{a}}}
\def\vb{{\bm{b}}}
\def\vc{{\bm{c}}}
\def\vd{{\bm{d}}}
\def\ve{{\bm{e}}}
\def\vf{{\bm{f}}}
\def\vg{{\bm{g}}}
\def\vh{{\bm{h}}}
\def\vi{{\bm{i}}}
\def\vj{{\bm{j}}}
\def\vk{{\bm{k}}}
\def\vl{{\bm{l}}}
\def\vm{{\bm{m}}}
\def\vn{{\bm{n}}}
\def\vo{{\bm{o}}}
\def\vp{{\bm{p}}}
\def\vq{{\bm{q}}}
\def\vr{{\bm{r}}}
\def\vs{{\bm{s}}}
\def\vt{{\bm{t}}}
\def\vu{{\bm{u}}}
\def\vv{{\bm{v}}}
\def\vw{{\bm{w}}}
\def\vx{{\bm{x}}}
\def\vy{{\bm{y}}}
\def\vz{{\bm{z}}}
\def\vpi{{\bm{\pi}}}

\def\evalpha{{\alpha}}
\def\evbeta{{\beta}}
\def\evepsilon{{\epsilon}}
\def\evlambda{{\lambda}}
\def\evomega{{\omega}}
\def\evmu{{\mu}}
\def\evpsi{{\psi}}
\def\evsigma{{\sigma}}
\def\evtheta{{\theta}}
\def\eva{{a}}
\def\evb{{b}}
\def\evc{{c}}
\def\evd{{d}}
\def\eve{{e}}
\def\evf{{f}}
\def\evg{{g}}
\def\evh{{h}}
\def\evi{{i}}
\def\evj{{j}}
\def\evk{{k}}
\def\evl{{l}}
\def\evm{{m}}
\def\evn{{n}}
\def\evo{{o}}
\def\evp{{p}}
\def\evq{{q}}
\def\evr{{r}}
\def\evs{{s}}
\def\evt{{t}}
\def\evu{{u}}
\def\evv{{v}}
\def\evw{{w}}
\def\evx{{x}}
\def\evy{{y}}
\def\evz{{z}}

\def\mA{{\bm{A}}}
\def\mB{{\bm{B}}}
\def\mC{{\bm{C}}}
\def\mD{{\bm{D}}}
\def\mE{{\bm{E}}}
\def\mF{{\bm{F}}}
\def\mG{{\bm{G}}}
\def\mH{{\bm{H}}}
\def\mI{{\bm{I}}}
\def\mJ{{\bm{J}}}
\def\mK{{\bm{K}}}
\def\mL{{\bm{L}}}
\def\mM{{\bm{M}}}
\def\mN{{\bm{N}}}
\def\mO{{\bm{O}}}
\def\mP{{\bm{P}}}
\def\mQ{{\bm{Q}}}
\def\mR{{\bm{R}}}
\def\mS{{\bm{S}}}
\def\mT{{\bm{T}}}
\def\mU{{\bm{U}}}
\def\mV{{\bm{V}}}
\def\mW{{\bm{W}}}
\def\mX{{\bm{X}}}
\def\mY{{\bm{Y}}}
\def\mZ{{\bm{Z}}}
\def\mBeta{{\bm{\beta}}}
\def\mPhi{{\bm{\Phi}}}
\def\mLambda{{\bm{\Lambda}}}
\def\mSigma{{\bm{\Sigma}}}
\def\lapsym{\mL_{\text{sym}}}
\def\laprw{\mL_{\text{rw}}}

\newcommand{\tens}[1]{\bm{\mathsfit{#1}}}
\def\tA{{\tens{A}}}
\def\tB{{\tens{B}}}
\def\tC{{\tens{C}}}
\def\tD{{\tens{D}}}
\def\tE{{\tens{E}}}
\def\tF{{\tens{F}}}
\def\tG{{\tens{G}}}
\def\tH{{\tens{H}}}
\def\tI{{\tens{I}}}
\def\tJ{{\tens{J}}}
\def\tK{{\tens{K}}}
\def\tL{{\tens{L}}}
\def\tM{{\tens{M}}}
\def\tN{{\tens{N}}}
\def\tO{{\tens{O}}}
\def\tP{{\tens{P}}}
\def\tQ{{\tens{Q}}}
\def\tR{{\tens{R}}}
\def\tS{{\tens{S}}}
\def\tT{{\tens{T}}}
\def\tU{{\tens{U}}}
\def\tV{{\tens{V}}}
\def\tW{{\tens{W}}}
\def\tX{{\tens{X}}}
\def\tY{{\tens{Y}}}
\def\tZ{{\tens{Z}}}

\def\gA{{\mathcal{A}}}
\def\gB{{\mathcal{B}}}
\def\gC{{\mathcal{C}}}
\def\gD{{\mathcal{D}}}
\def\gE{{\mathcal{E}}}
\def\gF{{\mathcal{F}}}
\def\gG{{\mathcal{G}}}
\def\gH{{\mathcal{H}}}
\def\gI{{\mathcal{I}}}
\def\gJ{{\mathcal{J}}}
\def\gK{{\mathcal{K}}}
\def\gL{{\mathcal{L}}}
\def\gM{{\mathcal{M}}}
\def\gN{{\mathcal{N}}}
\def\gO{{\mathcal{O}}}
\def\gP{{\mathcal{P}}}
\def\gQ{{\mathcal{Q}}}
\def\gR{{\mathcal{R}}}
\def\gS{{\mathcal{S}}}
\def\gT{{\mathcal{T}}}
\def\gU{{\mathcal{U}}}
\def\gV{{\mathcal{V}}}
\def\gW{{\mathcal{W}}}
\def\gX{{\mathcal{X}}}
\def\gY{{\mathcal{Y}}}
\def\gZ{{\mathcal{Z}}}

\def\sA{{\mathbb{A}}}
\def\sB{{\mathbb{B}}}
\def\sC{{\mathbb{C}}}
\def\sD{{\mathbb{D}}}
\def\sF{{\mathbb{F}}}
\def\sG{{\mathbb{G}}}
\def\sH{{\mathbb{H}}}
\def\sI{{\mathbb{I}}}
\def\sJ{{\mathbb{J}}}
\def\sK{{\mathbb{K}}}
\def\sL{{\mathbb{L}}}
\def\sM{{\mathbb{M}}}
\def\sN{{\mathbb{N}}}
\def\sO{{\mathbb{O}}}
\def\sP{{\mathbb{P}}}
\def\sQ{{\mathbb{Q}}}
\def\sR{{\mathbb{R}}}
\def\sS{{\mathbb{S}}}
\def\sT{{\mathbb{T}}}
\def\sU{{\mathbb{U}}}
\def\sV{{\mathbb{V}}}
\def\sW{{\mathbb{W}}}
\def\sX{{\mathbb{X}}}
\def\sY{{\mathbb{Y}}}
\def\sZ{{\mathbb{Z}}}

\def\emLambda{{\Lambda}}
\def\emA{{A}}
\def\emB{{B}}
\def\emC{{C}}
\def\emD{{D}}
\def\emE{{E}}
\def\emF{{F}}
\def\emG{{G}}
\def\emH{{H}}
\def\emI{{I}}
\def\emJ{{J}}
\def\emK{{K}}
\def\emL{{L}}
\def\emM{{M}}
\def\emN{{N}}
\def\emO{{O}}
\def\emP{{P}}
\def\emQ{{Q}}
\def\emR{{R}}
\def\emS{{S}}
\def\emT{{T}}
\def\emU{{U}}
\def\emV{{V}}
\def\emW{{W}}
\def\emX{{X}}
\def\emY{{Y}}
\def\emZ{{Z}}
\def\emSigma{{\Sigma}}

\newcommand{\etens}[1]{\mathsfit{#1}}
\def\etLambda{{\etens{\Lambda}}}
\def\etA{{\etens{A}}}
\def\etB{{\etens{B}}}
\def\etC{{\etens{C}}}
\def\etD{{\etens{D}}}
\def\etE{{\etens{E}}}
\def\etF{{\etens{F}}}
\def\etG{{\etens{G}}}
\def\etH{{\etens{H}}}
\def\etI{{\etens{I}}}
\def\etJ{{\etens{J}}}
\def\etK{{\etens{K}}}
\def\etL{{\etens{L}}}
\def\etM{{\etens{M}}}
\def\etN{{\etens{N}}}
\def\etO{{\etens{O}}}
\def\etP{{\etens{P}}}
\def\etQ{{\etens{Q}}}
\def\etR{{\etens{R}}}
\def\etS{{\etens{S}}}
\def\etT{{\etens{T}}}
\def\etU{{\etens{U}}}
\def\etV{{\etens{V}}}
\def\etW{{\etens{W}}}
\def\etX{{\etens{X}}}
\def\etY{{\etens{Y}}}
\def\etZ{{\etens{Z}}}

\newcommand{\pdata}{p_{\rm{data}}}
\newcommand{\ptrain}{\hat{p}_{\rm{data}}}
\newcommand{\Ptrain}{\hat{P}_{\rm{data}}}
\newcommand{\pmodel}{p_{\rm{model}}}
\newcommand{\Pmodel}{P_{\rm{model}}}
\newcommand{\ptildemodel}{\tilde{p}_{\rm{model}}}
\newcommand{\pencode}{p_{\rm{encoder}}}
\newcommand{\pdecode}{p_{\rm{decoder}}}
\newcommand{\precons}{p_{\rm{reconstruct}}}

\newcommand{\E}{\mathbb{E}}
\newcommand{\Ls}{\mathcal{L}}
\newcommand{\R}{\mathbb{R}}
\newcommand{\emp}{\tilde{p}}
\newcommand{\lr}{\alpha}
\newcommand{\reg}{\lambda}
\newcommand{\rect}{\mathrm{rectifier}}
\newcommand{\softmax}{\mathrm{softmax}}
\newcommand{\relu}{\mathrm{ReLU}}
\newcommand{\sigmoid}{\sigma}
\newcommand{\softplus}{\zeta}
\newcommand{\KL}{D_{\mathrm{KL}}}
\newcommand{\Var}{\mathrm{Var}}
\newcommand{\standarderror}{\mathrm{SE}}
\newcommand{\Cov}{\mathrm{Cov}}
\newcommand{\normlzero}{L^0}
\newcommand{\normlone}{L^1}
\newcommand{\normltwo}{L^2}
\newcommand{\normlp}{L^p}
\newcommand{\normmax}{L^\infty}

\newcommand{\parents}{Pa} 

\let\ab\allowbreak

\title{\model: Learning a Unified Cross-Domain Foundation Model for Text-Attributed Graphs}

\author{Yufei He}
\affiliation{National University of Singapore\country{Singapore}}
\email{yufei.he@u.nus.edu}

\author{Yuan Sui}
\affiliation{National University of Singapore\country{Singapore}}
\email{yuan.sui@u.nus.edu}

\author{Xiaoxin He}
\affiliation{National University of Singapore\country{Singapore}}
\email{he.xiaoxin@u.nus.edu}

\author{Bryan Hooi}
\affiliation{National University of Singapore\country{Singapore}}
\email{bhooi@comp.nus.edu.sg}

\begin{abstract}
Foundation models like ChatGPT and GPT-4 have revolutionized artificial intelligence, exhibiting remarkable abilities to generalize across a wide array of tasks and applications beyond their initial training objectives. However, graph learning has predominantly focused on single-graph models, tailored to specific tasks or datasets, lacking the ability to transfer learned knowledge to different domains. This limitation stems from the inherent complexity and diversity of graph structures, along with the different feature and label spaces specific to graph data. In this paper, we recognize text as an effective unifying medium and employ Text-Attributed Graphs (TAGs) to leverage this potential. We present our \model\footnote[1]{The code is available at \url{https://github.com/yf-he/UniGraph}} framework, designed to learn a foundation model for TAGs, which is capable of generalizing to unseen graphs and tasks across diverse domains. Unlike single-graph models that use pre-computed node features of varying dimensions as input, our approach leverages textual features for unifying node representations, even for graphs such as molecular graphs that do not naturally have textual features. We propose a novel cascaded architecture of Language Models (LMs) and Graph Neural Networks (GNNs) as backbone networks. Additionally, we propose the first pre-training algorithm specifically designed for large-scale self-supervised learning on TAGs, based on Masked Graph Modeling. We introduce graph instruction tuning using Large Language Models (LLMs) to enable zero-shot prediction ability. Our comprehensive experiments across various graph learning tasks and domains demonstrate the model's effectiveness in self-supervised representation learning on unseen graphs, few-shot in-context transfer, and zero-shot transfer, even surpassing or matching the performance of GNNs that have undergone supervised training on target datasets.
\end{abstract}



\maketitle
\section{Introduction}
Foundation models in artificial intelligence are large-scale pre-trained models that provide a versatile base, enabling a wide range of applications and tasks~\cite{bommasani2021opportunities}.
However, in the graph learning community, the long-standing practice is to train a model specific to one graph at a time~\cite{kipf2016semi}, which we call a single-graph model.
These models often can only handle one or a few tasks and lack the ability to transfer to other graphs.
Also, single-graph models typically require a substantial amount of labeled data for each specific task, which can be a significant limitation in data-scarce scenarios.

\vpara{Challenges in building a graph foundation model.}
The success of language and vision foundation models is built upon invariances across their respective application domains, such as a unified vocabulary or raw pixels.
The major challenge in learning a graph foundation model is the diversity of graph domains and, furthermore, how to learn the invariances across different domains.
Firstly, graphs from different domains have distinct feature spaces and label spaces. In NLP, texts from different domains, despite having vastly different semantics, can still be encoded using the same dictionary, thereby generating transferable representations.
However, for graphs from different domains, their nodes and edges have different types and semantics, leading to incompatibility between their features, making it very difficult to unify model inputs.
Furthermore, graph models like GNNs~\cite{kipf2016semi}, use a static softmax classifier for prediction, lacking the capability for zero-shot prediction on unseen classes in different label spaces.

Secondly, as a universal data structure applied across various domains, graphs exhibit significant structural differences.
Citation networks are directed and often acyclic, with nodes representing scholarly articles and edges representing citations. In contrast, knowledge graphs are more complex, with nodes representing entities and diverse edges representing relations, often forming intricate patterns like cycles and cliques.
It is challenging for graph models to transfer learned structural knowledge across different domains.

\vpara{Presented work.}
Motivated by these challenges, in this work, 
we propose the \model framework to design and train a foundation model capable of generalizing to a variety of tasks on unseen graphs across diverse domains.

To align the feature spaces of different graphs, we use text as a unifying medium.
Many graphs in the real world contain textual features, such as citation networks and Wikipedia networks, and are known as Text-Attributed Graphs (TAGs). 
Compared to pre-processed vector features, textual features provide a consistent representation across different domains, which is beneficial for the transferability of the model.
In addition to natural TAGs, we also explore using text to represent features of non-natural TAGs, such as molecular graphs, making it possible for the model to generalize to a broader range of domains.

To learn transferable invariances across different graphs and tasks, we design a universal template to unify different tasks by contextualizing the nodes/edges/graphs on which we make predictions. 
Some perspectives suggest that representative local graph structural patterns are universal and transferable across different graphs~\cite{newman2018networks}.
For node/edge-level tasks on large-scale graphs, we adopt the Personalized PageRank (PPR) algorithm to sample subgraphs, thereby mitigating differences in graph structures across different domains, while also making the model scalable.

We propose a cascaded architecture of LMs and GNNs and new pre-training objectives based on Masked Graph Modeling (MGM), specifically tailored for self-supervised learning on TAGs at scale.
To further endow our model with zero-shot capabilities, we concatenate pre-trained graph embeddings with natural language instructions and perform instruction tuning on LLMs, allowing us to unify the label spaces of graphs from different domains through natural language in a generative manner.

To summarize, our work makes the following contributions:
\begin{itemize}[leftmargin=*,itemsep=0pt,parsep=0.2em,topsep=0.3em,partopsep=0.3em]
    \item 
    We identify the challenges in developing a cross-domain graph foundation model and present the use of TAGs to unify graphs from diverse domains. We propose a foundation model for TAGs, named \model, which incorporates a novel cascaded LM and GNN backbone. Additionally, we propose the first pre-training algorithm specifically designed for large-scale self-supervised learning on TAGs.
    \item We explore the use of graph instruction tuning to leverage the powerful generalization capabilities of LLMs for making zero-shot predictions on graph learning tasks.
    \item We conduct experiments on node/edge/graph-level tasks across 11 different graph datasets from 5 distinct domains, with the largest graph comprising 111 million nodes. In cross-domain settings, \model outperforms not just other cross-domain methods, but also supervised methods trained on the target dataset.
\end{itemize}

\section{Related Work}
\vpara{Cross-graph learning.}
For graph transfer learning within the same domain, the most successful are pre-trained models on molecular graphs~\cite{liu2021pre}, benefiting from similar node/edge semantics. 
In addition, techniques such as fine-tuning~\cite{hou2022graphmae,he2024generalizing}, domain adaptation~\cite{dai2022graph}, and prompt graphs~\cite{huang2023prodigy} are used to achieve cross-graph transfer within the same domain. 
However, all these methods require the alignment of different graphs in both vector feature and label space.
For cross-domain graph learning,
GCC~\cite{qiu2020gcc} and ULTRA~\cite{galkin2023towards}, by ignoring node features and pretraining structural representations, achieve transfer learning for specific tasks, such as knowledge graph reasoning. 
OFA~\cite{liu2023one} leverages pre-trained LMs to align the feature spaces of TAGs from different domains, while also transforming all downstream classification tasks into binary classification tasks, enabling it to conduct supervised training across all graphs.

\vpara{Connections to existing methods.}
As a foundation model for TAGs, our \model framework is distinct from OFA in three main aspects: 1) \model performs end-to-end representation learning on TAGs, whereas OFA's training is decoupled, using frozen LMs to generate features and then training GNNs separately. 
2) \model is a self-supervised learning framework, whereas OFA is a supervised learning framework that requires specific downstream task labels.
3) \model, after pretraining, can generalize to any unseen target TAGs. In contrast, OFA co-trains on multiple target graphs simultaneously and then infers on each separately.
Another type of existing graph learning models on TAGs are LLMs-only methods like LLaGA~\cite{chen2024llaga} and GraphGPT~\cite{tang2023graphgpt}, which use instruction tuning to map graph data into the LLM embedding space. UniGraph can integrate with these methods, offering high-quality embeddings that improve LLMs' understanding of graph structures.
For conventional LMs+GNNs models that are not cross-domain models, they differ from UniGraph in both the target research problem and technical details. For example, GLEM~\cite{zhao2022learning} and TAPE~\cite{he2023harnessing} are supervised learning models on TAGs; they decouple LMs and GNNs, optimizing them separately. G2P2~\cite{wen2023augmenting} uses a CLIP~\cite{radford2021learning}-like contrastive learning algorithm and employs prompt tuning to adapt to different downstream tasks.

\begin{figure*}[t]
    \centering
    \includegraphics[width=1\linewidth]{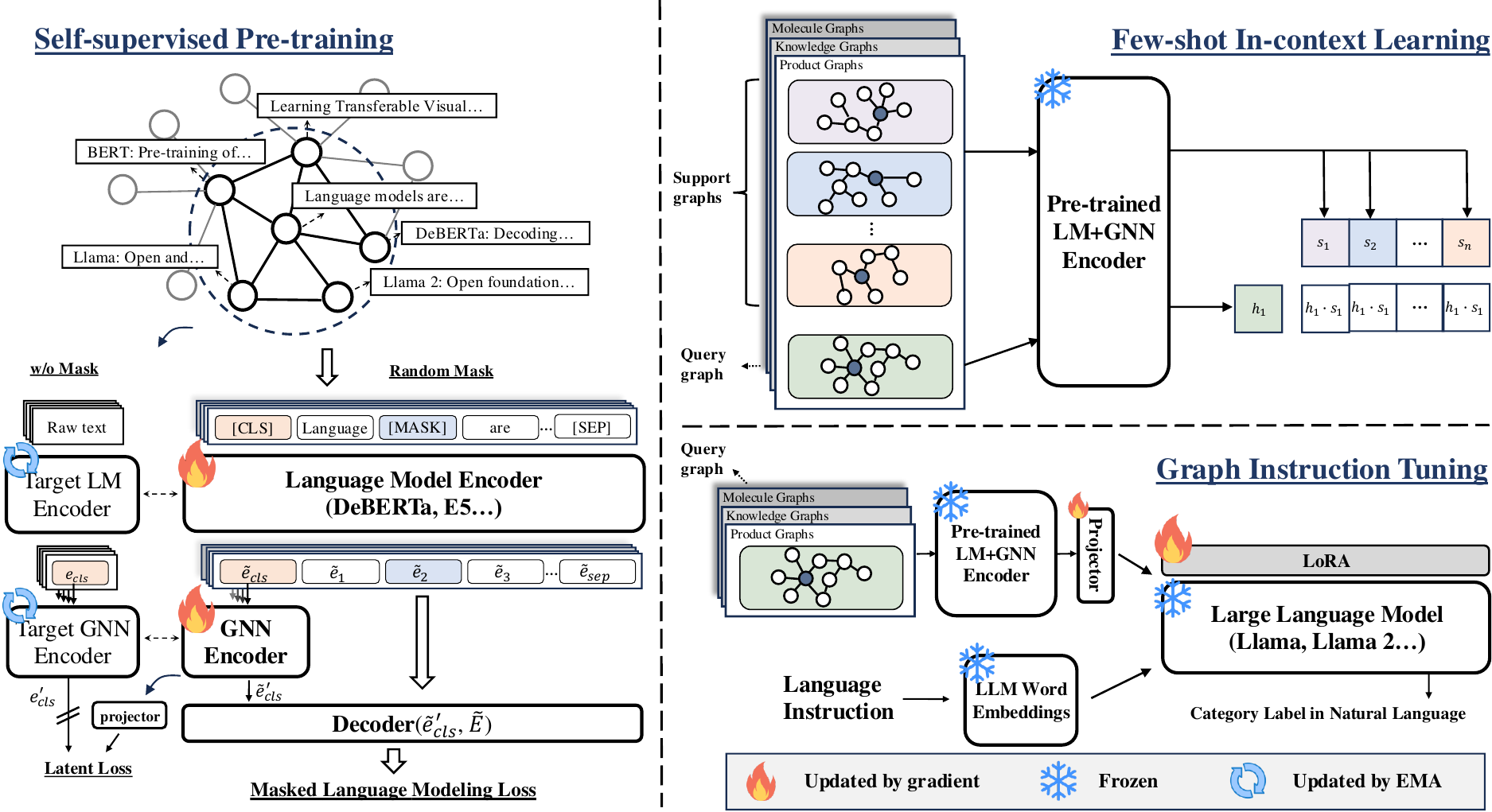}
    \vskip 0.0in
    \caption{\textbf{Overview of \model framework.}
    1) In pre-training, we employ a self-supervised approach, leveraging TAGs to unify diverse graph data. This phase involves a cascaded architecture combining LMs and GNNs. 
    We propose Graph Siamese Masked Autoencoders as the training architecture, which learns to reconstruct the masked text of each node using the text of its neighbors.
    2) In few-shot transfer, the pre-trained model can make predictions with minimal data by comparing the embeddings of the query and support graphs. 
    3) Zero-shot transfer is achieved through graph instruction tuning with LLMs, enabling it to understand category labels in natural language and make predictions on unseen graphs without any graph-specific training.}
    \vspace{-3mm}
    \label{fig:arch}
\end{figure*}

\section{Preliminaries}
\subsection{Text-Attributed Graphs (TAGs)}
\begin{definition}
\label{def:tags}
A Text-Attributed Graph (TAG) is defined as a graph \( \mathcal{G} = (\mathcal{V}, \mathcal{E}, \mathcal{T}_\mathcal{V}, \mathcal{T}_\mathcal{E}) \), where \( \mathcal{V} \) represents the set of nodes and \( \mathcal{E} \) represents the set of edges. 
For each node \( v \in \mathcal{V} \), there is an associated text \( t_{v} \in \mathcal{T}_\mathcal{V} \) representing node-level textual information. 
For each edge \( e_{vu} \in \mathcal{E} \) connecting nodes \( v \) and \( u \), there is an associated text \( t_{e_{vu}} \in \mathcal{T}_\mathcal{E} \) representing edge-level textual information. 
It is possible for a TAG to have only \( \mathcal{T}_\mathcal{V} \).
\end{definition}
\subsection{Problem Definition}
\label{sec:ssl}
This paper introduces a cross-domain foundation model for TAGs, pre-trained in a self-supervised manner. 
Our model's generalization ability is evaluated across domains using unseen datasets through three machine learning problems.
The pre-trained model, denoted as \(f_{\theta}\), 
operates on a TAG and generates an embedding for each node. Formally, the model's function can be expressed as:
$f_{\theta}: \mathcal{G} \rightarrow \mathbb{R}^{|V| \times d}$,
where \(d\) is the dimensionality of the embedding space.

\vpara{Self-supervised representation learning.} 
The primary aim of self-supervised learning is to produce embeddings that are useful in downstream tasks. 
We adopt a linear probing setting to evaluate the representation learning ability of a frozen pre-trained model \( f_{\theta} \). 

\vpara{Few-shot transfer.} 
In the Few-Shot Transfer problem, 
\(N\)-way \(K\)-shot tasks evaluate the model's in-context learning ability to apply its pre-learned knowledge to a new task with $N$ classes, each represented by only $K$ labeled examples. 

\vpara{Zero-shot transfer.} 
In \(N\)-way \(K\)-shot tasks, \(K\) is set to 0, indicating no prior exposure to support samples from the target classes. 
This setting is to evaluate a pre-trained model's ability to generalize and apply its learned knowledge to unseen data categories. 

\section{The \model Approach}
In this section, we present the \model framework, as illustrated in Figure~\ref{fig:arch}.
\subsection{Unifying Graphs and Tasks from Diverse Domains}
Graphs from different domains often have different applications, corresponding to different tasks.
Graph learning tasks can generally be divided into node, edge, and graph-level tasks, each focusing on different parts of the graph.
The key to using one model to handle any task on any graph lies in finding a universal function acting as a versatile mapping tool, adaptable to different graph learning tasks. 
In this paper, we utilize the concept of Anchor Node(s) and their contextual subgraph(s) to construct this universal function.
The unification of node, edge, and graph-level tasks can be achieved through a general   
contextual subgraphs processing and Anchor Nodes embedding refinement operation, denoted by $\mathbf{h} = g(\mathcal{X}, \mathcal{A})$, where \( \mathbf{h} \) represents the output vector representation for a node, an edge or a graph, \( \mathcal{X} \) denotes the set of contextual subgraphs, and \( \mathcal{A} \) signifies the set of Anchor Node(s).
Formally, we define the function \( g \) with our pre-trained model \( f_{\theta} \) using the format: 
\begin{equation}
    \mathbf{h} = g(\mathcal{X}, \mathcal{A}) = R(f_{\theta}(\mathcal{X}), \mathcal{A}),
\end{equation}
\( R \) is the task-specific readout function that aggregates the output of \( f_{\theta} \) and produces the final vector representation \( \mathbf{h} \).

\vpara{Node-level.}
The Anchor Node in node-level task is a single node \( v \) for which we aim to generate the embedding, \( \mathcal{X} = \{\mathcal{G}_v\} \) and \( \mathcal{A} = \{v\} \). $R_{\text{node}}$ simply extracts its embedding.

\vpara{Edge-level.}
The Anchor Nodes in edge-level tasks are the nodes \( v \) and \( u \) forming an edge, \( \mathcal{X} = \{\mathcal{G}_v, \mathcal{G}_u\} \) and \( \mathcal{A} = \{v, u\} \). 
$R_{\text{edge}}$ extracts their corresponding embeddings and concatenates them.

\vpara{Graph-level.}
In graph-level tasks, since the graphs for graph-level tasks are often smaller, all nodes in \( \mathcal{G} \) are Anchor Nodes, \( \mathcal{X} = \{\mathcal{G}\} \) and \( \mathcal{A} = \mathcal{V} \). \( R_{\text{graph}} \) is a pooling function such as average pooling.

\vpara{PPR for subgraph sampling.}
For a given Anchor Node \( v \) on a graph \( \mathcal{G} = (\mathcal{V}, \mathcal{E}) \), we utilize the top-k Personalized PageRank algorithm to sample its contextual subgraph for node and edge-level tasks. 
The PPR algorithm computes a relevance score for each node in \( \mathcal{G} \) with respect to the Anchor Node \( v \).
The contextual subgraph \( \mathcal{G}_v \) is then composed of the top-k nodes with the highest PPR scores, including their connecting edges. 
Formally, \( \mathcal{C}_v \) is the set of top-k nodes based on PPR scores, then:
 $\mathcal{G}_v = (\mathcal{V}_v, \mathcal{E}_v) \text{ where } \mathcal{V}_v = \{v\} \cup \mathcal{C}_v \text{ and } \mathcal{E}_v = \{(u, w) \in \mathcal{E} : u, w \in \mathcal{V}_v \}$.

By adopting Personalized PageRank (PPR) as the sampling strategy, we can construct the most structurally relevant~\cite{bianchini2005inside} and information-rich~\cite{gasteiger2018predict} local subgraphs for Anchor Nodes, while also enabling our model to scale to web-scale graphs.
Compared to other local sampling strategies like k-hop neighbors, PPR can identify crucial nodes and structures that are important in a broader context, which can be more universally transferable~\cite{lofgren2016personalized}.

\vspace{-2mm}
\subsection{Graph Siamese Masked Autoencoders on TAGs}
The existing graph self-supervised learning methods~\cite{velivckovic2018deep,kipf2016variational,thakoor2021bootstrapped,hou2022graphmae} adopt GNNs as the backbone networks and use pre-processed vector features as input.
In this part, we propose an end-to-end self-supervised learning method on TAGs. We cascade a pre-trained language model (LM) and a GNN as the backbone network. 
In this paper, we adopt DeBERTa-base~\cite{he2020deberta} as our LM and GAT~\cite{velivckovic2018graph} as our GNN.
Inspired by successful self-supervised learning techniques such as masked modeling~\cite{kenton2019bert,he2022masked,hou2022graphmae,zhao2024pre} and siamese networks~\cite{wu2018unsupervised}, we designed Graph Siamese Masked Autoencoders, enabling large-scale self-supervised pre-training on TAGs.

\vpara{Masking.}
In the training, a (sub)graph \( \mathcal{G} = (\mathcal{V}, \mathcal{E}, \mathcal{T}_\mathcal{V}) \) is processed in each batch, where \( \mathcal{V} \) represents the set of nodes, \( \mathcal{E} \) the set of edges, and \( \mathcal{T}_\mathcal{V} \) the textual features associated with each node. The textual features for each node are extended with a [CLS] token at the beginning and a [SEP] token at the end. The [CLS] token is treated as the embedding of the sentence/node.
Let \( t_v \) denote the textual feature sequence for node \( v \in \mathcal{V} \), extended with [CLS] and [SEP] tokens. The sequence for each node is then tokenized into a sequence of tokens \( t_v = [\text{[CLS]}, T_{1}, T_{2}, \ldots, T_{n_v}, \text{[SEP]}] \), where \( n_v \) is the number of tokens for node \( v \).
The masking process involves randomly replacing a subset of tokens in each \( t_v \) with a [MASK] token, defined by the masking function: 
\begin{equation}
    m_v = \text{Mask}(t_v) = [\text{[CLS]}, M_{1}, M_{2}, \ldots, M_{n_v}, \text{[SEP]}],
\end{equation}
where \( M_{i} = \begin{cases} \text{[MASK]}, & \text{with probability } p \\ T_{i}, & \text{otherwise}. \end{cases} \)

\vpara{Encoder.}
Our encoder consists of an LM \( f_{\text{LM}} \) and a GNN \( f_{\text{GNN}} \).
For each node \( v \in \mathcal{V} \), the masked textual feature sequence \( m_v \) is processed through the LM \( f_{\text{LM}} \) to get the hidden representations: \( \widetilde{\mE}_v = f_{\text{LM}}(m_v) \). 
Then, we extract embedding of the [CLS] token $\widetilde{\ve}_{v_\text{cls}} \in \mathbb{R}^{d}$ from the masked output $\widetilde{\mE}_v \in \mathbb{R}^{(n_v+2) \times d}$. 
The GNN \( f_{\text{GNN}} \) propagates embeddings of the [CLS] tokens across the graph. 
The input to \( f_{\text{GNN}} \) is a matrix \(\widetilde{\mE}_{\text{cls}} \in \mathbb{R}^{|\mathcal{V}| \times d} \) consisting of embedding of the [CLS] token for all nodes.
The output of the GNN, \( \widetilde{\mE}'_{\text{cls}} \in \mathbb{R}^{|\mathcal{V}| \times d} \), is another matrix of embeddings, representing the propagated features: \( \widetilde{\mE}'_{\text{cls}} = f_{\text{GNN}}(\mathcal{G}, \widetilde{\mE}_{\text{cls}}) \)

\vpara{Decoder.}
In decoding, we adopt the masked language modeling (MLM)~\cite{kenton2019bert} as objective.
The intuition behind designing such training objectives is that the model can learn to reconstruct the masked text of each node using the text of its neighbors, thereby fully exploring the graph structure while learning to understand the text.
For each node \( v \), the embeddings of masked textual feature $\widetilde{\mE}_v$ from the masked forward pass and the GNN output of [CLS] token $\widetilde{\ve}'_{v_{\text{cls}}}$ are concatenated and linearly transformed:
\begin{equation}
    \mH_v = \text{Linear}(\widetilde{\mE}_v \oplus (\widetilde{\ve}'_{v_{\text{cls}}}\otimes \mathbf{1}_{n_v+2}^\top)),
\end{equation}
where \( \mathbf{1}_{n_v+2} \) is a column vector of ones with a length of \( n_v+2 \). \( \otimes \) in this context represents the outer product, which replicates the vector \( \widetilde{\ve}'_{v_{\text{cls}}} \) to form a matrix whose number of rows matches the number of tokens in \( \widetilde{\mE}_v \). 
The resulting matrix of $\widetilde{\ve}'_{v_{\text{cls}}}\otimes \mathbf{1}_{n_v+2}^\top$ has same dimensions as $\widetilde{\mE}_v \in \mathbb{R}^{(n_v+2) \times d}$.
The concatenation operation \(\oplus\) in this context is the horizontal joining of two matrices, taking two matrices of dimensions \((n_v+2) \times d\) each and resulting in a single matrix of dimensions \((n_v+2) \times 2d\).
The linear function can be expressed as: $\text{Linear}(*): \mathbb{R}^{2d} \rightarrow \mathbb{R}^{d}$.
Then, we use an MLMHead, which is an MLP, to map transformed embeddings to vocabulary space, producing probability distributions: $\mathbf{P}_v = \text{MLMhead}(\mathbf{H}_v)$. 

The training loss is computed using a CrossEntropy loss function for each node \( v \) on the graph, aiming to predict the original tokens at the masked positions:
\begin{equation}
    \mathcal{L}_{\text{mask}} = -\frac{1}{\sum_v \sum_i \mathbb{I}(v, i)} \sum_{v \in \mathcal{V}}\sum_{i=1}^{n_v} \mathbb{I}(v, i) \cdot \log\mathbf{P}_v[i,T_i]
\end{equation}
where $\mathbb{I}(v, i)$ is an indicator function that is 1 if the \( i \)-th token in the sentence/node \( v \) is a [MASK] token and 0 otherwise. 
$\mathbf{P}_v[i,T_i]$ refers to the probability assigned to the true token $T_i$ for the $i$-th token in the sentence/node $v$.

\begin{table*}[t]
\centering
\caption{\textbf{Statistics of all eleven text-attributed graph datasets.}}
\vskip -0.1in
\label{tab:dataset}
\begin{tabular}{lrrrrrc}
\toprule[1.1pt]
Dataset & Domain & Task & \#Graphs & Avg. \#Nodes & Avg. \#Edges & Raw Texts \\
\midrule
Cora & Citation & Node & 1 & 2,708 & 5,429  & paper titles and abstracts \\
PubMed & Citation & Node & 1 & 19,717 & 44,338 & paper titles and abstracts \\
ogbn-Arxiv & Citation & Node & 1 & 169,343 & 1,166,243  & paper titles and abstracts \\
ogbn-Papers100M & Citation & Node & 1 & 111,059,956	& 1,615,685,872	& paper titles and abstracts \\
ogbn-Products & Product  & Node & 1 & 2,449,029 & 61,859,140 & product descriptions \\
Wiki-CS & Web link & Node & 1 & 11,701 & 216,123 & wikipedia entry names and contents \\
FB15K237 & Knowledge & Edge & 1 & 14,541 & 310,116 &  entity names and descriptions \\
WN18RR & Knowledge & Edge & 1 & 40,943 & 93,003 &  entity names and descriptions \\
PCBA & Molecule & Graph & 437,929 & 26.0 & 28.1  & textual descriptions of atoms/bonds\\
HIV & Molecule & Graph & 41,127 & 25.5 & 27.5 & textual descriptions of atoms/bonds \\
ChEMBL & Molecule & Graph & 365,065 & 25.9 & 55.9 & textual descriptions of atoms/bonds \\
\bottomrule[1.1pt]
\end{tabular}
\end{table*}

\begin{table*}[t]
    \centering
    \renewcommand\tabcolsep{4.8pt}
    \caption{\textbf{Experiment results in self-supervised representation learning.} We report accuracy (\%) for node/edge classification tasks and ROC-AUC score (\%) for graph classification tasks. The performance of supervised methods, which are trained on the individual target dataset, is marked in \colorbox{Gray}{gray}. \model and other self-supervised baselines (rows in white) are pretrained on ogbn-Papers100M, and then evaluated on the individual target dataset. NoPretrain represents a randomly initialized model in our framework without any pre-training.
    }
    \vskip -0.10in
    \label{tab:ssrl}
    \begin{tabular}{lccccccccc}
    \toprule[1.1pt]
    & \multicolumn{5}{c}{Node Classification} & \multicolumn{2}{c}{Edge Classification}& \multicolumn{2}{c}{Graph Classification}\\
    \cmidrule(lr){2-6}\cmidrule(lr){7-8}\cmidrule(lr){9-10}
         & Cora & PubMed & Arxiv & Products & Wiki-CS & FB15K237 & WN18RR & HIV & PCBA \\
    \midrule 
    \multicolumn{5}{l}{\textbf{Use word2vec to encode raw text as input features.}} \\
    Linear  & 50.12{\small$\pm$0.12} & 61.99{\small$\pm$0.21} & 50.11{\small$\pm$0.17} & 66.29{\small$\pm$0.21} & 66.23{\small$\pm$0.11} & 81.21{\small$\pm$0.21} & 69.03{\small$\pm$0.32} & 60.99{\small$\pm$0.31} & 54.35{\small$\pm$1.34}\\
    DGI  & 51.99{\small$\pm$0.45} & 55.76{\small$\pm$0.56} & 55.21{\small$\pm$0.21} & 64.21{\small$\pm$0.32} & 67.11{\small$\pm$0.12} & 26.99{\small$\pm$0.22} & 52.04{\small$\pm$0.22} & 60.12{\small$\pm$0.32} & 54.22{\small$\pm$1.23}\\
    BGRL   & 56.73{\small$\pm$0.23} & 63.77{\small$\pm$0.23} & 62.21{\small$\pm$0.21} & 66.22{\small$\pm$0.39} & 70.12{\small$\pm$0.15} & 64.91{\small$\pm$0.22} & 56.44{\small$\pm$0.21} & 60.67{\small$\pm$0.39} & 54.89{\small$\pm$1.11}\\
    GraphMAE   & 60.12{\small$\pm$0.87} & 66.22{\small$\pm$0.35} & 65.22{\small$\pm$0.22} & 67.19{\small$\pm$0.39} & 68.11{\small$\pm$0.12} & 61.11{\small$\pm$0.12} & 59.76{\small$\pm$0.29} & 59.21{\small$\pm$0.31} & 52.10{\small$\pm$1.24}\\
    GraphMAE2  & 61.19{\small$\pm$0.45} & 65.99{\small$\pm$0.21} & 67.19{\small$\pm$0.11} & 67.73{\small$\pm$0.12} & 68.84{\small$\pm$0.37} & 63.76{\small$\pm$0.12} & 60.24{\small$\pm$0.23} & 60.23{\small$\pm$0.35} & 53.90{\small$\pm$0.99}\\
    \rowcolor{Gray} GCN  & 71.98{\small$\pm$1.33} & 69.86{\small$\pm$1.01} & 70.11{\small$\pm$0.14} & 79.12{\small$\pm$0.12} & 78.12{\small$\pm$0.37} & \underline{90.21{\small$\pm$0.56}} & 74.21{\small$\pm$0.63} & 70.11{\small$\pm$1.35} & \textbf{60.23{\small$\pm$0.45}}\\
    \rowcolor{Gray} GAT  & 72.42{\small$\pm$1.21} & \underline{70.45{\small$\pm$1.21}} & 70.89{\small$\pm$0.43} & \underline{79.67{\small$\pm$0.34}} & \underline{79.09{\small$\pm$0.67}} & 88.65{\small$\pm$0.26} & \underline{74.80{\small$\pm$0.64}} & \underline{71.12{\small$\pm$1.34}} & 56.24{\small$\pm$1.01}\\
    \midrule
    \multicolumn{5}{l}{\textbf{Use DeBERTa-base to encode raw text as input features.}} \\
    Linear & 29.34{\small$\pm$0.11} & 48.51{\small$\pm$0.22} & 43.22{\small$\pm$0.25} & 41.29{\small$\pm$0.21} & 41.09{\small$\pm$0.10} & 78.11{\small$\pm$0.32} & 65.03{\small$\pm$0.11} & 60.11{\small$\pm$0.34} & 53.46{\small$\pm$1.02}\\
    DGI  & 30.36{\small$\pm$0.36} & 52.91{\small$\pm$0.51} & 49.15{\small$\pm$0.21} & 56.18{\small$\pm$0.36} & 63.15{\small$\pm$0.52} & 29.12{\small$\pm$0.13} & 51.98{\small$\pm$0.53} & 59.12{\small$\pm$0.34} & 53.23{\small$\pm$0.47}\\
    BGRL   & 40.10{\small$\pm$0.34} & 52.99{\small$\pm$0.41} & 56.19{\small$\pm$0.22} & 60.15{\small$\pm$0.44} & 66.87{\small$\pm$0.32} & 45.69{\small$\pm$0.25} & 46.15{\small$\pm$0.39} & 61.33{\small$\pm$0.62} & 54.22{\small$\pm$1.04}\\
    GraphMAE    & 43.11{\small$\pm$0.51} & 54.14{\small$\pm$0.32} & 57.11{\small$\pm$0.64} & 65.22{\small$\pm$0.43} & 69.01{\small$\pm$0.33} & 56.21{\small$\pm$0.21} & 53.22{\small$\pm$0.39} & 62.01{\small$\pm$0.65} & 51.45{\small$\pm$1.01}\\
    GraphMAE2   & 42.87{\small$\pm$0.43} & 53.98{\small$\pm$0.31} & 59.39{\small$\pm$0.49} & 67.91{\small$\pm$0.48} & 70.47{\small$\pm$0.13} & 55.82{\small$\pm$0.28} & 51.78{\small$\pm$0.24} & 61.42{\small$\pm$0.61} & 52.35{\small$\pm$0.35}\\
    \rowcolor{Gray} GCN  & 48.42{\small$\pm$1.33} & 60.33{\small$\pm$1.98} & 60.76{\small$\pm$2.42} & 66.98{\small$\pm$2.32}  & 77.43{\small$\pm$0.43} & 85.23{\small$\pm$0.65} & 72.04{\small$\pm$0.32} & 66.24{\small$\pm$1.31} & 58.21{\small$\pm$1.04}\\
    \rowcolor{Gray} GAT  & 47.99{\small$\pm$1.89} & 61.01{\small$\pm$1.18} & 63.11{\small$\pm$2.24} & 67.02{\small$\pm$2.11}  & 78.10{\small$\pm$0.34} & 83.01{\small$\pm$1.01} & 73.98{\small$\pm$0.23} & 67.12{\small$\pm$1.23} & 56.45{\small$\pm$0.45}\\
    \midrule
    \multicolumn{5}{l}{\textbf{Use raw text as input features.}} \\
    GIANT-XRT  & 70.23{\small$\pm$0.87} & 64.35{\small$\pm$0.43} & 70.87{\small$\pm$0.11} & 66.93{\small$\pm$0.32} & 70.13{\small$\pm$0.88} & 89.65{\small$\pm$0.85} & 72.78{\small$\pm$0.66} & 65.14{\small$\pm$0.32} & 51.34{\small$\pm$1.98}\\
    +GraphMAE2 & \underline{80.11{\small$\pm$0.35}} & 69.43{\small$\pm$0.45} & \underline{72.01{\small$\pm$0.24}} & 75.23{\small$\pm$0.34} & 76.58{\small$\pm$0.21} & 76.12{\small$\pm$1.03} & 57.32{\small$\pm$0.66} & 67.23{\small$\pm$0.98} & 52.01{\small$\pm$0.45}\\
    \hdashline
    NoPretrain & 40.98{\small$\pm$0.32} & 53.01{\small$\pm$0.35} & 62.22{\small$\pm$0.20} & 67.12{\small$\pm$0.21} & 73.21{\small$\pm$0.15} & 23.19{\small$\pm$0.21} & 51.03{\small$\pm$0.29} & 58.01{\small$\pm$0.21} & 51.01{\small$\pm$0.43}\\
    \model & \textbf{81.43{\small$\pm$0.55}} & \textbf{74.33{\small$\pm$0.23}} & \textbf{72.91{\small$\pm$0.42}} & \textbf{80.11{\small$\pm$0.23}}  & \textbf{79.98{\small$\pm$1.21}} & \textbf{94.81{\small$\pm$1.32}} & \textbf{85.45{\small$\pm$0.34}} & \textbf{71.23{\small$\pm$1.93}} & \underline{57.67{\small$\pm$0.85}}\\
    \bottomrule[1.1pt]
    \end{tabular}
    \vspace{-3mm}
\end{table*}

\vpara{Latent space regularization.}
Some recent works suggest that designing new pretext tasks in the latent space helps stabilize the training of masked autoencoders, further enhancing the quality of the latent space representations~\cite{chen2023context,dong2022bootstrapped,zhang2021scr,he2022sgkd,zhang2021improving}.
In this work, we propose the use of a target network to impose regularization constraints on the latent space. 
The target network shares the same architecture as the encoder, comprising an LM \( f'_{\text{LM}} \) and a GNN \( f'_{\text{GNN}} \), with parameters not updated through gradients.
The parameters of \( f'_{\text{LM}}(*;\delta') \) and \( f_{\text{LM}}(*;\delta) \) are shared, while the parameters of \( f'_{\text{GNN}}(*;\xi') \) are updated through exponential moving average (EMA) of \( f_{\text{GNN}}(*;\xi) \) using weight decay $\tau$ to avoid collapse in negative-free SSL frameworks~\cite{grill2020bootstrap, caron2021emerging}: $\delta' \leftarrow  \delta \quad \text{and} \quad \xi' \leftarrow  \tau \xi' + (1-\tau)\xi.$

During training, \( f'_{\text{LM}} \) processes the original, unmasked textual feature \( t_v \):\( \mE_v = f_{\text{LM}}(t_v). \) 
We feed the embeddings of [CLS] tokens for all nodes \( \mE_{\text{cls}} \) into the \( f'_{\text{GNN}} \) to get the propagated embeddings \( \mE'_{\text{cls}} \). 
Then the encoding results \( \widetilde{\mE}_{\text{cls}} \) of the masked graph are projected to representation space by a linear projector, resulting in $\bar{\mZ}$ for latent regularization with a latent loss:
\begin{equation}
    \mathcal{L}_{\text{latent}} = \frac{1}{|\mathcal{V}|} \sum_{i}^{|\mathcal{V}|} (1 - \frac{\bar{\vz}_i^\top \ve'_{{\text{cls}}_i} }{\lVert \bar{\vz}_i \rVert \cdot \lVert \ve'_{{\text{cls}}_i} \rVert })
\end{equation}
The encoder and projector network are trained to match the output of the target network. 

Then we obtain the overall loss by fusing the two losses with a mixing coefficient $\lambda$:
\begin{equation}
    \mathcal{L} = \mathcal{L}_{\text{mask}} + \lambda \cdot \mathcal{L}_{\text{latent}}.
\end{equation}
\vpara{Inference.}
In inference, we discard the decoder and target network, using only the encoder to generate embeddings. Given an unseen (sub)graph \( \mathcal{G'} = (\mathcal{V'}, \mathcal{E'}, \mathcal{T}_\mathcal{V'}) \) and the Anchor Nodes for which we want to obtain embeddings, we first utilize \( f_{\text{LM}} \) to encode the raw text of all nodes on the graph.
Subsequently, we feed the [CLS] tokens of all nodes into \( f_{\text{GNN}} \) to obtain the propagated representations, which serve as the final embeddings of each node. Finally, we extract the embeddings corresponding to the Anchor Nodes.

\vpara{Enabling in-context learning.}
\label{sec:icl}
A dataset after generating embeddings, comprising node/edge/graph embedding-label pairs \((\vh, y)\). 
An $N$-way $K$-shot transfer task involves a support set \(\mathcal{S} = \{(\vh_i, y_i)\}_{i=1}^{N \times K}\) and a query set \(\mathcal{Q}\). 
For each class \(c\), the model computes an average embedding \(\overline{\ve_c} = \frac{1}{K} \sum_{(\vh, y) \in \mathcal{S}, y = c} \vh\). 
The classification of a query data sample \(\vh_q\) in \(\mathcal{Q}\) is determined by comparing its embedding to these average embeddings using the cosine similarity. 
The performance of the model is evaluated using a metric \(\mathcal{M}\), which could be accuracy, defined as:
\begin{equation}
    \mathcal{M} = \frac{1}{|\mathcal{Q}|} \sum_{\vh_q \in \mathcal{Q}} \mathbbm{1}\left[\arg\max_{c \in \{1, \ldots, N\}} \frac{\vh_q \cdot \overline{e_c}}{\|\vh_q\| \|\overline{e_c}\|} = y_q\right].
\end{equation}
\begin{table}[t]
    \centering
    \renewcommand\tabcolsep{2.8pt}
    \caption{\textbf{Experiment results in few-shot transfer.} We report accuracy (\%) for both datasets. Performance of OFA, which is directly trained on test datasets with labels, is marked in \colorbox{Gray}{gray}. Prodigy is pre-trained on the MAG240M or Wiki.
    \model and other self-supervised learning baselines are pre-trained on ogbn-Papers100M, and then evaluated on individual target dataset.}
    \vskip -0.1in
    \label{tab:fst}
    \begin{tabular}{lcccccc}
    \toprule[1.1pt]
    & \multicolumn{3}{c}{Arxiv} & \multicolumn{3}{c}{FB15K237} \\
    \cmidrule(lr){2-4}\cmidrule(lr){5-7}
         & 40-way & 5-way & 3-way & 40-way & 10-way & 5-way \\
    \midrule 
    \multicolumn{7}{l}{\textbf{Use word2vec to encode raw text as input features.}} \\
    word2vec &14.42 & 43.24 & 54.24 & 33.93& 68.24 & 75.80 \\
     DGI   &15.67 & 46.12 & 57.33 & 31.67& 67.75 & 74.26 \\
     BGRL  &17.98 & 48.43 & 60.24 & 29.24& 67.23 & 74.14 \\
     GraphMAE &19.12 & 49.24 & 62.34 & 32.07 & 69.75 & 77.24 \\
     GraphMAE2 &18.45 & 50.01 & 61.35 & 33.01& 70.45 & 78.01 \\
    \rowcolor{Gray}GPN & 7.24 & 34.53 & 49.23 & - & - & - \\
    \rowcolor{Gray}TENT & 10.34 & 41.24 & 60.98 & - & - & - \\
    \rowcolor{Gray}GLITTER & 14.23 & 44.24 & 60.21 & - & - & - \\
    \midrule 
    \multicolumn{7}{l}{\textbf{Use LMs to encode raw text as input features.}} \\
    Prodigy &25.13 &61.52 & 73.09 &  59.58 &81.1 & 88.02 \\
    \rowcolor{Gray}OFA & 22.13 & 60.12 &  72.17 &\underline{65.23} & \underline{83.01} &  \underline{90.11} \\
    \rowcolor{Gray}GPF+ & 18.01 & 56.12 & 65.23 & 60.12 & 80.12 & 86.23 \\
    \rowcolor{Gray}All in One & 19.87 & 57.24 & 71.34 & 62.24 & 69.23 & 87.24 \\
    \rowcolor{Gray}GraphPrompt & 16.23 & 59.23 & 72.19 & 66.32 & 67.21 & 85.23 \\
    \midrule 
    \multicolumn{7}{l}{\textbf{Use raw text as input features.}} \\
    GIANT-XRT & 20.12 & 54.33 & 59.98 & 52.63 & 77.21 & 85.57 \\
    +GraphMAE2 & \underline{27.35} & \underline{66.91} & \underline{74.62} & 47.73 & 74.33 & 80.17 \\
    \hdashline 
    NoPretrain  & 12.57 & 39.46 & 49.16 & 27.39 & 62.94 & 74.84 \\
    \model  & \textbf{31.35} & \textbf{74.12} & \textbf{83.24} & \textbf{68.76} & \textbf{85.32} & \textbf{91.12} \\
    \bottomrule[1.1pt]
    \end{tabular}
\end{table}

\begin{table*}[t]
    \centering
    \caption{\textbf{Experiment results in zero-shot transfer.} We report accuracy (\%) for all datasets. \model-IT represents our self-supervised pre-trained model with our instruction tuned LLM.}
    \vskip -0.1in
    \label{tab:zs}
    \begin{threeparttable}
    \begin{tabular}{lccccccccccc}
    \toprule[1.1pt]
    & \multicolumn{2}{c}{Cora} & \multicolumn{1}{c}{PubMed} & \multicolumn{3}{c}{Products$^1$} & \multicolumn{2}{c}{Wiki-CS} & \multicolumn{3}{c}{WN18RR} \\
    \cmidrule(lr){2-3}\cmidrule(lr){4-4}\cmidrule(lr){5-7}\cmidrule(lr){8-9}\cmidrule(lr){10-12}
    & 7-way & 2-way &  3-way & 47-way & 10-way &5-way & 10-way & 5-way &11-way &10-way & 5-way \\
    \midrule
    Llama-7B & 33.43 & 57.32 & 46.33 & 13.45 & 32.53 & 40.24 & 15.32 & 26.32 & 12.53 & 13.56 & 27.21 \\
    Llama2-7B & 34.21 & 58.99 & 43.57 & 16.53 & 35.25 & 42.29 & 20.34 & 31.24 & 14.21 & 14.98 & 29.29  \\
    vicuna-7B-v1.5 & 45.23 & 72.21 & 62.14 & 20.24 & 41.23 & 54.45 & 29.46 & 45.21 & 23.24 & 26.24 & 34.14\\
    \midrule 
    OFA  & 24.01 & 56.92 & 54.01 & - & - & - & - & - & 18.43 & 19.98 & 30.96 \\
    GraphGPT & - & - & \underline{70.11} & - & - & - & - & - & - & - & - \\
    G2P2$^2$  & 54.29 & 82.41 & 68.23 & 30.82 & 48.20 & 70.91 & 30.11 & 47.12 & 26.24& 27.42 & 42.14 \\
    ZeroG$^2$ & \underline{64.21} & \underline{87.83} & 67.23 & \underline{31.24} & \underline{51.24} & \underline{71.29} & \underline{31.26} & \underline{48.25} & \underline{31.42} & \underline{30.21} & \underline{43.28}\\
    \midrule 
    \model-IT  & \textbf{69.53} & \textbf{89.74} & \textbf{72.48} & \textbf{38.45} & \textbf{66.07} & \textbf{75.73} & \textbf{43.45} & \textbf{60.23} & \textbf{36.73} & \textbf{38.24} & \textbf{54.32} \\
    \bottomrule[1.1pt]
    \end{tabular}
    \begin{tablenotes}
            \footnotesize
            \item \tiny The results of open-source models not reported are due to being unavailable in their papers or source code. 
            \item[1] \tiny Since the test set of Products is quite large, we randomly sample 50,000 nodes from it for evaluation. 
            \item[2] \tiny Due to the difference between the pre-trained datasets used in their original paper and ours, we re-run the experiments using their open-source code.
        \end{tablenotes}
    \end{threeparttable}
    \vspace{-2.3mm}
\end{table*}

\subsection{Graph Instruction Tuning}
Graphs from different domains have distinct label spaces, making it challenging to directly transfer to unseen graphs without fine-tuning the pre-trained graph models using a substantial number of labels~\cite{hou2023graphmae2}.
To enhance zero-shot capabilities, we propose using a graph instruction tuned open-ended generative LLM to unify the label spaces of different graphs. In instruction tuning, we provide textual instructions to the model which vary for different domains, to facilitate adaptation to each domain. To train the model to use these instructions, we fine-tune the model on instruction tuning datasets where we have labels.

\vpara{Instruction prompts.}
For instruction tuning, we design prompt templates that include graph embeddings, graph structure, and natural language instructions, as summarized in Appendix~\ref{appendix:prompts}. 

\vpara{Training.} 
Our instruction tuning pipeline is shown in Figure~\ref{fig:arch}. 
Given a query graph, we first generate embeddings for it using our pretrained model \(f_{\theta}\).
We apply a linear projector to map the graph embeddings into the LLM's embedding space. 
Then we combine graph embeddings and natural language instructions as inputs to the LLM.
We select Llama-7B~\cite{touvron2023llama1} as our LLM and adopt LoRA~\cite{hu2021lora} for fine-tuning the LLM while keeping the word embeddings frozen.
We fine-tune the LLM to generate labels in natural language and the loss is computed only on the predicted target.


\section{Experiments}
In this section, we evaluate the cross-domain generalization ability of our \model framework on three distinct research problems. Table~\ref{tab:dataset} shows all the datasets in the experiments.

\subsection{Self-Supervised Representation Learning}
\label{sec:lp}
\vpara{Setup.}
We adopt the most commonly used linear probing protocol to evaluate the representation learning ability of self-supervised pre-trained models on unseen datasets. We train a linear classifier on top of the embeddings from a frozen model.
Our model and all self-supervised learning baseline methods are first pre-trained on the large-scale citation network ogbn-Papers100M~\cite{hu2020open}. Then, we evaluate on nine graphs from five domains with different tasks.
Detailed settings and hyper-parameters can be found in Appendix~\ref{appendix:imple}.

For the baselines, we compare \model with state-of-the-art generative graph self-supervised learning methods: GraphMAE~\cite{hou2022graphmae} and GraphMAE2~\cite{hou2023graphmae2}, constrastive methods: DGI~\cite{velivckovic2018deep} and BGRL~\cite{thakoor2021bootstrapped}. 
As they are not designed for cross-domain purposes, we utilize shallow LM word2vec~\cite{mikolov2013distributed} and pre-trained LM DeBERTa-base~\cite{he2020deberta} to unify the input node features of different graphs. 
Consistent with our approach, all baseline methods use GAT~\cite{velivckovic2018graph} as the backbone GNN.
We also compare with supervised models GCN~\cite{kipf2016semi} and GAT, which are separately trained on the target datasets. 
For baseline methods that take TAGs as input, we select GIANT-XRT, which fine-tunes language models using graph structure, and GIANT-XRT+GraphMAE2, which conducts second pre-training based on embeddings generated by GIANT-XRT.

\vpara{Results.}
Table~\ref{tab:ssrl} presents the results.
We interpret these results from three perspectives:
(1) \model outperforms state-of-the-art graph self-supervised learning methods by a large margin. 
This indicates that our framework possesses a stronger generalization ability in cross-domain graph learning scenarios, enabling it to generate more discriminative embeddings for unseen graphs.
(2) Compared to using pre-processed features, learning directly from TAGs is more advantageous for cross-domain transfer. \model and GIANT-XRT, which take text as input, demonstrate stronger performance than GNN-based methods that use LMs to pre-encode text features. 
(3) As a single pre-trained model applied to various downstream datasets, \model exhibits better or comparable performance than supervised learning models trained directly on those downstream datasets. This further suggests the feasibility and effectiveness of training a self-supervised graph foundation model.

\subsection{Few-Shot Transfer}
\label{sec:fs}

\vpara{Setup.}
In this part, we evaluate the ability of the pre-trained models to perform few-shot in-context transfer without updating the model parameters. 
For baseline methods, in addition to the pre-trained models mentioned in section~\ref{sec:lp}, we also compared two latest graph in-context learning methods: the self-supervised pre-training method Prodigy~\cite{huang2023prodigy} and the supervised pre-training method OFA~\cite{liu2023one}.
They each utilize different LMs to unify the input features of different graphs.
Unlike our setting, Prodigy is pre-trained on the MAG240M~\cite{hu2021ogb} or Wiki datasets~\cite{huang2023prodigy} for corresponding downstream Arxiv or FB15K237 datasets, while OFA is pre-trained on Arxiv, FB15K237, and ChEMBL.
We also compare \model with graph prompt learning methods, such as GPF \cite{fang2024universal}, All in One \cite{sun2023all}, and GraphPrompt \cite{liu2023graphprompt}. Additionally, we compare UniGraph with graph meta-learning methods, including GPN \cite{ding2020graph}, TENT \cite{wang2022task}, and GLITTER \cite{wang2022graph}.

For evaluation, we strictly follow the setting of Prodigy~\cite{huang2023prodigy}. 
For an N-way K-shot task, we adopt the original train/validation/test splits in each downstream classification dataset, and construct a $K$-shot prompt for test nodes (or edges) from the test split by randomly selecting $K$ examples per way from the train split. By default in all experiments, we sample 500 test tasks with 3-shot prompts.

\vpara{Results.}
In table~\ref{tab:fst}, the results demonstrate that our \model framework consistently outperforms all the baselines. 
In particular, compared to Prodigy and OFA, which are pre-trained on the same tasks as the downstream tasks, our model still achieves superior performance.
This suggests that our graph foundation model is capable of learning effective general knowledge from pre-training tasks and can learn in the context of downstream tasks.  
For graph prompt learning and graph meta learning methods, we can observe that although these baseline methods are fine-tuned on downstream datasets, UniGraph is still able to outperform these baselines without modifying the model parameters.

\subsection{Zero-Shot Transfer}
\vpara{Setup.}
For zero-shot transfer, we mainly compare with general-purpose LLMs such as Llama~\cite{touvron2023llama1}, Llama2~\cite{touvron2023llama2} and vicuna-7B-v1.5~\cite{vicuna2023}, as well as graph-based LLM methods
GraphGPT~\cite{tang2023graphgpt}, ZeroG~\cite{li2024zerog} and G2P2~\cite{wen2023augmenting}. 
Also, we compare with OFA mentioned in section~\ref{sec:fs}.
Consistent with OFA, we select ogbn-Arxiv and FB15K237 as instruction tuning datasets. 
Note that there is no overlap between their label categories and those of the downstream datasets. We report the average accuracy on the original test set with 3 random initializations for each downstream dataset. 

\vpara{Results.}
In table~\ref{tab:zs}, we can observe that our proposed \model significantly outperforms open-source LLMs, confirming that our framework can effectively align graph embeddings with natural language representations. Also, LLMs are capable of learning transferable graph structure and graph learning task knowledge from our designed graph instruction tuning. Additionally, compared to other graph zero-shot learning methods, UniGraph also demonstrates consistent advantages.

\begin{table*}[t]\small
\centering
\caption{\textbf{Performance comparison with dataset-specific graph self-supervised learning.}}
\vskip -0.1in
\label{tab:single}
\renewcommand\tabcolsep{3.3pt}
\begin{tabular}{lccccccccccc}
\toprule[1.1pt]
Method & Cora & PubMed & Arxiv & Products & Papers100M& Wiki-CS & FB15K237 & WN18RR & HIV & PCBA & ChEMBL\\ 
\midrule
DGI-single & 69.24\tiny$\pm$0.42 & 68.24\tiny$\pm$0.24 & 69.76\tiny$\pm$0.33 & 78.11\tiny$\pm$0.24 & 55.21\tiny$\pm$0.49 & 77.76\tiny$\pm$0.24 & 85.27\tiny$\pm$0.64 & 52.04\tiny$\pm$0.22 & 68.42\tiny$\pm$0.90 & 51.24\tiny$\pm$0.87 & 61.24\tiny$\pm$0.53\\ 
BGRL-single & 71.24\tiny$\pm$0.11 & 69.01\tiny$\pm$0.34 & 70.87\tiny$\pm$0.24 & 79.01\tiny$\pm$0.32 & 64.45\tiny$\pm$0.32 & 78.24\tiny$\pm$0.24 & 87.24\tiny$\pm$0.25 & 72.11\tiny$\pm$0.68 & 68.23\tiny$\pm$0.33 & 55.82\tiny$\pm$1.23 & 62.04\tiny$\pm$0.54 \\ 
GraphMAE-single & 73.24\tiny$\pm$0.76 & 69.23\tiny$\pm$0.97 & 72.01\tiny$\pm$0.24 & 78.24\tiny$\pm$0.37 & 65.43\tiny$\pm$0.76& 79.24\tiny$\pm$0.22 & 86.35\tiny$\pm$0.35 & 72.34\tiny$\pm$0.24 & 69.23\tiny$\pm$0.23 & 56.22\tiny$\pm$1.87 & 63.28\tiny$\pm$0.76\\ 
GraphMAE2-single & 72.23\tiny$\pm$0.33 & 69.87\tiny$\pm$0.43 & 71.21\tiny$\pm$0.24 & 79.11\tiny$\pm$0.24 & 66.24\tiny$\pm$0.54& 79.21\tiny$\pm$0.34 & 87.22\tiny$\pm$0.24 & 70.45\tiny$\pm$0.24 & 68.34\tiny$\pm$0.34 & 56.87\tiny$\pm$0.43 & 63.54\tiny$\pm$0.73\\ 
\midrule
UniGraph-cross & \underline{81.43\tiny$\pm$0.55} & \underline{74.33\tiny$\pm$0.23} & \underline{72.91\tiny$\pm$0.42} & \underline{80.11\tiny$\pm$0.23} & \textbf{67.89\tiny$\pm$0.21} & \underline{79.98\tiny$\pm$1.21} & \underline{94.81\tiny$\pm$1.32} & \underline{85.45\tiny$\pm$0.34} & \underline{71.23\tiny$\pm$1.93} & \underline{57.67\tiny$\pm$0.85} & \underline{64.29\tiny$\pm$1.01} \\ 
UniGraph-single & \textbf{84.23\tiny$\pm$0.24} & \textbf{80.11\tiny$\pm$0.21} & \textbf{73.97\tiny$\pm$0.22} & \textbf{82.24\tiny$\pm$0.24} & \textbf{67.89\tiny$\pm$0.21} & \textbf{81.22\tiny$\pm$0.24} & \textbf{95.24\tiny$\pm$0.23} & \textbf{87.21\tiny$\pm$0.76} & \textbf{75.24\tiny$\pm$1.24} & \textbf{60.23\tiny$\pm$0.11} & \textbf{65.32\tiny$\pm$0.21}\\ 
\bottomrule[1.1pt]
\end{tabular}

\end{table*}

\begin{table*}[t]
\centering
\caption{\textbf{Comparison of computational costs and performance on ogbn-Arxiv and ogbn-Papers100M.}}
\vskip -0.1in
\label{tab:ccp}
\begin{tabular}{lccccc}
\toprule[1.1pt]
Dataset & Method & Pre-training Time & Downstream Training Time & Downstream Inference Time & Test Accuracy \\
\midrule
\multirow{3}{*}{\shortstack{ogbn-Arxiv \\ (169,343 nodes)}} 
  & GAT        & -    & 23.2 mins & 5.5 mins  & 70.89 $\pm$ 0.43 \\
  & GraphMAE2  & -    & 4.8 h     & 5.1 mins  & 71.21 $\pm$ 0.24 \\
  & UniGraph   & 23.4 h & -      & 10.2 mins & 72.91 $\pm$ 0.42 \\
\midrule
\multirow{3}{*}{\shortstack{ogbn-Papers100M \\ (111,059,956 nodes)}}
  & GAT        & -    & 6.8 h     & 23.1 mins & 65.98 $\pm$ 0.23 \\
  & GraphMAE2  & -    & 20.1 h    & 24.3 mins & 66.24 $\pm$ 0.54 \\
  & UniGraph   & 23.4 h & -      & 41.2 mins & 67.89 $\pm$ 0.21 \\
\bottomrule[1.1pt]
\end{tabular}
\vspace{-2.3mm}
\end{table*}

\begin{table}[t]
    \centering
    \renewcommand\tabcolsep{3.5pt}
    \caption{\textbf{Ablation studies of \model key components.}}
    \vskip -0.1in
    \label{tab:kc}
    \begin{tabular}{lcccc}
    \toprule[1.1pt]
        & Arxiv & Products & WN18RR & HIV \\
    \midrule
        \model & 72.91{\footnotesize$\pm$0.42} & 80.11{\footnotesize$\pm$0.23} & 85.45{\footnotesize$\pm$0.34} & 71.23{\footnotesize$\pm$1.93}\\
        w/o GNN & 68.24{\footnotesize$\pm$0.52} & 64.24{\footnotesize$\pm$0.66} & 76.24{\footnotesize$\pm$0.24} & 56.11{\footnotesize$\pm$1.18}\\
        w/o MLM loss& 67.86{\footnotesize$\pm$0.52} & 68.53{\footnotesize$\pm$0.91} & 78.22{\footnotesize$\pm$0.21} & 53.25{\footnotesize$\pm$1.01}\\
        w/o latent loss& 72.24{\footnotesize$\pm$0.24} & 78.99{\footnotesize$\pm$0.90} & 84.72{\footnotesize$\pm$0.32} & 70.53{\footnotesize$\pm$1.34}\\
        w/o PPR sampling & 72.01{\footnotesize$\pm$0.21} & 79.23{\footnotesize$\pm$0.54} & 83.53{\footnotesize$\pm$0.31} & 70.64{\footnotesize$\pm$1.01}\\
    \bottomrule[1.1pt]
    \end{tabular}
    \vspace{-2mm}
\end{table}

\begin{table}[t]
    \centering
    \renewcommand\tabcolsep{3.5pt}
    \caption{Ablation studies of \model pre-training datasets. 'citation', 'KG', and 'MOL' respectively refer to ogbn-Papers100M, FB15K237, and ChEMBL.}
    \vskip -0.1in
    \label{tab:pd}
    \begin{tabular}{lcccc}
    \toprule[1.1pt]
        & Arxiv & Products & WN18RR & HIV \\
    \midrule
        citation & 72.91{\footnotesize$\pm$0.42} & 80.11{\footnotesize$\pm$0.23} & 85.45{\footnotesize$\pm$0.34} & 71.23{\footnotesize$\pm$1.93}\\
        citation+MOL & 72.83{\footnotesize$\pm$0.55} & 80.01{\footnotesize$\pm$0.21} & 84.37{\footnotesize$\pm$0.24} & 77.84{\footnotesize$\pm$1.01}\\
        citation+KG & 72.96{\footnotesize$\pm$0.57} & 79.36{\footnotesize$\pm$0.75} & 91.01{\footnotesize$\pm$0.34} & 69.24{\footnotesize$\pm$1.12}\\
       citation+KG+MOL & 72.78{\footnotesize$\pm$0.23} & 79.53{\footnotesize$\pm$0.74} & 83.25{\footnotesize$\pm$0.22} & 75.43{\footnotesize$\pm$0.86}\\
    \bottomrule[1.1pt]
    \end{tabular}
    \vspace{-2mm}
\end{table}

\begin{table}[t]
    \centering
    \renewcommand\tabcolsep{3.5pt}
    \caption{\textbf{Analysis of LMs and GNNs choices.}}
    \vskip -0.1in
    \label{tab:lm}
    \begin{tabular}{lcccc}
    \toprule[1.1pt]
        & Arxiv & Products & WN18RR & HIV \\
    \midrule
        DeBERTa-base & 72.91{\footnotesize$\pm$0.42} & 80.11{\footnotesize$\pm$0.23} & 85.45{\footnotesize$\pm$0.34} & 71.23{\footnotesize$\pm$1.93}\\
        DeBERTa-v3-base & 72.73{\footnotesize$\pm$0.53} & 80.34{\footnotesize$\pm$0.21} & 84.99{\footnotesize$\pm$0.55} & 70.87{\footnotesize$\pm$1.76}\\
        DeBERTa-large & 73.21{\footnotesize$\pm$0.55} & 81.24{\footnotesize$\pm$0.21} & 85.99{\footnotesize$\pm$0.55} & 71.68{\footnotesize$\pm$1.54}\\
        E5-large-v2 & 73.19{\footnotesize$\pm$0.35} & 81.27{\footnotesize$\pm$0.24} & 86.21{\footnotesize$\pm$0.31} & 71.56{\footnotesize$\pm$1.22}\\
    \midrule
        GAT & 72.91{\footnotesize$\pm$0.42} & 80.11{\footnotesize$\pm$0.23} & 85.45{\footnotesize$\pm$0.34} & 71.23{\footnotesize$\pm$1.93}\\
        GCN & 72.25{\footnotesize$\pm$0.34} & 79.43{\footnotesize$\pm$0.35} & 86.11{\footnotesize$\pm$0.35} & 72.86{\footnotesize$\pm$1.21}\\
    \bottomrule[1.1pt]
    \end{tabular}
    \vspace{-2.0mm}
\end{table}

\begin{table}[t]
    \centering
    \caption{\textbf{Efficiency comparison of our method versus only using LM for pretraining on an NVIDIA A100 (40G) GPU.}}
    \vskip -0.1in
    \label{tab:ec}
    \begin{tabular}{lccc}
    \toprule[1.1pt]
        & \#parameters & speed & memory \\
    \midrule
        DeBERTa-base & 180,209,243 & 2.48it/s & 38397 MB\\
        \model & 181,984,093 & 2.43it/s & 38724 MB \\
    \bottomrule[1.1pt]
    \end{tabular}
    \vspace{-2.0mm}
\end{table}

\subsection{Comparisons With Dataset-Specific Graph Self-Supervised Learning}
Table~\ref{tab:single} shows the performance under self-supervised representation learning setting, where "-single" indicates self-supervised training on the target dataset, and "-cross" signifies self-supervised pre-training on the pre-training dataset Papers100M. For DGI, BGRL, GraphMAE, and GraphMAE2, word2vec features are used as model inputs, whereas for UniGraph, text is directly used as input.

From the experimental results, we observe that UniGraph still outperforms dataset-specific Graph SSL, further illustrating the potential and feasibility of foundation models. At the same time, the performance of UniGraph with dataset-specific self-supervised training exceeds that of cross-domain UniGraph to some extent. This indicates the effectiveness of the proposed SSL algorithm and suggests that there is still room for improvement in cross-domain pre-training. Note that cross-domain UniGraph avoids the need for collecting additional task-specific data for training.

\subsection{Model Analysis}
We choose four datasets from different domains to conduct more in-depth studies. We adopt self-supervised representation learning for evaluation.

\vpara{Ablation on key components.}
Table~\ref{tab:kc} shows the performance of the \model framework after removing some key designs. "W/o GNN" represents that we use standard MLM loss to finetune the LM. 
"W/o MLM loss" represents that we only use the latent loss for pre-training, while "w/o latent loss" refers to the opposite.
"W/o PPR sampling" refers to not using PPR for sampling and instead employing neighbor sampling.
The overall results confirm that all key designs contribute to the performance of \model.

\vpara{Ablation on pre-training datasets.}
Table~\ref{tab:pd} shows the impact of including graphs from different domains in the pre-training datasets on downstream performance.
We can observe that pre-training on graphs from the same domain enhances the performance of downstream tasks. This suggests that in-domain transfer remains simpler than cross-domain transfer.
However, this paper primarily focuses on the generalization ability of the graph foundation model in cross-domain transfer. 
For fair comparison with baselines that train only on a single dataset, we restrict our pre-training to ogbn-Papers100M.

\vpara{Analysis of LMs and GNNs choices.}
In table~\ref{tab:lm}, we study the influence of different choices of LMs and GNNs that constitute the backbone network on performance.
Compared to DeBERTa-base(50K), DeBERTa-V3-base has a larger dictionary size(128K).
DeBERTa-large has a larger backbone(350M) than DeBERTa-base(100M).
We also try E5-large-v2, which is a text embedding LM.
The results show that larger LMs can achieve better performance, but practical experiments may need to consider the trade-off between performance and speed.

\subsection{Efficiency Analysis.}
For \model pre-training, we denote the maximum sequence length of node textual feature as \(L\) and the number of nodes processed in each batch as \(N\).

\vpara{Time complexity.}
The time complexity for pre-training is dominated by the LM processing, which scales as \(O(N \cdot (L^2d + Ld^2))\), where \(d\) is the dimensionality of the embeddings. The GNN adds a complexity of \(O(N \cdot d^2)\) for immediate neighborhood aggregation, potentially increasing to \(O(N \cdot d^2 + N^2 \cdot d)\) for dense graphs. Latent loss calculation, such as cosine similarity, adds \(O(N \cdot d)\). Thus, the overall complexity is primarily driven by the LM with \(O(N \cdot (L^2d + Ld^2))\), with the GNN contributing a secondary, though not negligible, cost.

\vpara{Space complexity.}
The space complexity for the LM includes storage for intermediate activations and model parameters, amounting to \(O(N \cdot L \cdot d)\) for activations and \(O(N \cdot L^2)\) for the self-attention mechanism, leading to a total of \(O(N \cdot L^2 + N \cdot L \cdot d)\). For the GNN, the space requirement is primarily for storing node features, approximated as \(O(N^2 + d^2 + N \cdot d)\) for dense graphs. Therefore, the overall space complexity is dominated by the LM, with \(O(N \cdot L^2 + N \cdot L \cdot d)\), while the GNN adds a relatively smaller contribution.

\vpara{Pretraining efficiency.}
In Table~\ref{tab:ec}, we can observe that the time and space overhead of training \model is comparable to only training an LM with the MLM task.
We can conclude that the computation cost of our framework is dominated by LM cost, and our method's running time is similar to other LM-based methods if they use similar LMs. 

\vpara{Computational costs as a foundation model.}
\label{computaional}
\model, designed as a foundation model, incurs significant computational costs primarily during the pre-training phase. However, it offers the advantage of applicability to new datasets in the inference phase without necessitating retraining. We conduct comparisons of the training/inference costs between our model and GNN-based models. GAT~\cite{velivckovic2018graph} is a supervised trained GNN. GraphMAE2~\cite{hou2023graphmae2} is a self-supervised learning method with GAT as the backbone network. We choose ogbn-Arxiv and ogbn-Papers100M, two datasets of different scales for experiments. 
From the results in the table~\ref{tab:ccp}, we can observe that although UniGraph has a long pre-training time, its inference time on downstream datasets is comparable/shorter than the training plus inference time of GNN-based methods. This advantage further increases with the size and potential quantity of downstream datasets.
The same conclusion also applies to space complexity. Although LM has a larger number of parameters, since we only need to perform inference on the downstream dataset, we avoid the additional space occupation in the backward propagation during training. 
\section{Conclusion}
In this work, we present \model framework, aimed at designing and training a novel foundation model for TAGs that enables cross-domain generalization. 
We have demonstrated that through large-scale pre-training, our framework can effectively learn and transfer knowledge across diverse graph domains.
The experimental results, covering a wide range of graph learning tasks and scenarios, 
validate the robustness and versatility of \model. 
This work not only addresses a critical gap in graph learning but also lays down a foundational framework that can be further explored and refined for broader applications. 

\begin{acks}
    This research is supported by the Ministry of Education, Singapore, under the Academic Research Fund Tier 2 (FY2025) (Award MOE-T2EP20124-0009).
\end{acks}


\bibliographystyle{ACM-Reference-Format}
\balance
\bibliography{reference}

\appendix

\begin{table*}[t] 
    \centering
    \caption{Pre-training hyper-parameters for our framework on ogbn-Papers100M.}
    \vskip 0.1in
    \label{tab:hyper}
    \renewcommand\tabcolsep{2.8pt}
    \begin{tabular}{ccccccccccc}
    \toprule[1.1pt]
       mask rate  & hidden\_size & lr & weight\_decay & dropout & optimizer & num\_epochs & num\_gnn\_layers & ppr topk & ema decay & coefficient $\lambda$\\
    \midrule
      0.75  & 768 & 2e-5 & 0.001 & 0.2 & adamw & 1 & 3 & 128 & 0.996 & 0.1\\
    \bottomrule[1.1pt]
    \end{tabular}
\end{table*}

\begin{table*}[t] 
    \centering
    \caption{Graph instruction tuning hyper-parameters for our framework on ogbn-Arxiv and FB15K237.}
    \vskip 0.1in
    \label{tab:git}
    \renewcommand\tabcolsep{2.3pt}
    \begin{tabular}{ccccccccccc}
    \toprule[1.1pt]
       LLM  & hidden\_size & lr & weight\_decay & dropout & optimizer & num\_epochs & warmup\_ratio & clip\_grad\_norm & batch\_size & max\_text\_length\\
    \midrule
      Llama  & 4096 & 8e-5 & 0.0 & 0.0 & adamw & 2 & 0.05 & 1.0 & 4 & 1024\\
    \bottomrule[1.1pt]
    \end{tabular}
\end{table*}

\section{Datasets}

\vpara{Cora}~\cite{he2023harnessing}. The Cora dataset consists of 2708 scientific publications classified into one of seven classes – case based, genetic algorithms, neural networks, probabilistic methods, reinforcement learning, rule learning, and theory. The citation network consists of 5429 links. We collect raw text from ~\cite{he2023harnessing}.

\vpara{PubMed}~\cite{he2023harnessing}. The Pubmed dataset consists of 19,717 scientific publications from PubMed database pertaining to diabetes classified into one of three classes – Experimental induced diabetes, Type 1 diabetes, and Type 2 diabetes. 
As in~\cite{liu2023one}, we ask ChatGPT to generate a detailed description of each category. The citation network consists of 44,338 links. We collect raw text from ~\cite{he2023harnessing}.

\vpara{ogbn-Arxiv}~\cite{hu2020open}. The ogbn-arxiv dataset is a directed graph, representing the citation network between all Computer Science (CS) arXiv papers. Each node is an arXiv paper and each directed edge indicates that one paper cites another one. The task is to predict the 40 subject areas of arXiv CS papers, e.g.,, cs.AI, cs.LG, and cs.OS. We collect raw text from ~\cite{hu2020open}.

\vpara{ogbn-Papers100M}~\cite{hu2020open}. The ogbn-papers100M dataset is a directed citation graph of 111 million papers. We collect raw text from ~\cite{hu2020open}.

\vpara{ogbn-Products}~\cite{hu2020open}. The ogbn-products dataset is an undirected and unweighted graph, representing an Amazon product co-purchasing network. Nodes represent products sold in Amazon, and edges between two products indicate that the products are purchased together. The task is to predict the category of a product in a multi-class classification setup, where the 47 top-level categories are used for target labels. We collect raw text from ~\cite{hu2020open}.

\vpara{Wiki-CS}~\cite{liu2023one}. Wiki-CS is a Internet link network with each node represent a Wikipedia page and each edge represent the reference link. Each node’s label corresponds to the category of the entry. We collect raw text from ~\cite{liu2023one}.

\vpara{FB15K237}~\cite{liu2023one}. FB15K237 is a kowledge graph that contains knowledge base relation triples and textual mentions of Freebase entity pairs. We collect raw text from ~\cite{liu2023one}. Given that we propose a self-supervised learning framework, and the edge text features are the labels to be predicted, we solely utilized node text features and did not employ edge text features.

\vpara{WN18RR}~\cite{liu2023one}. WN18RR is a knowledge graph, which is a subset of WordNet that consists of 11 relations and 40943 entities.
We collect raw text from ~\cite{liu2023one}. Given that we propose a self-supervised learning framework, and the edge text features are the labels to be predicted, we solely utilized node text features and did not employ edge text features.

\vpara{PCBA}~\cite{liu2023one}. PCBA is a widely used molecule property prediction dataset. It contains 1,310 prediction target labels of molecules from biological assays for drug discovery. 
We collect raw text from ~\cite{liu2023one}.

\vpara{HIV}~\cite{liu2023one}. HIV is a subset of the BioChem BioAssay dataset consisting of 128 labels on the biological activities of small molecules.
We collect raw text from ~\cite{liu2023one}.

\vpara{ChEMBL}~\cite{liu2023one}. ChEMBL contains over 40,000 compounds labeled for their ability to inhibit HIV replication.
We collect raw text from ~\cite{liu2023one}.

Example for PCBA, HIV and ChEMBL:

Node textual features: \textit{atom. \textless element name \textgreater, \textless atom chirality \textgreater, degree of \textless atom degree \textgreater, formal charge of \textless formal charge \textgreater, num of hydrogen is \textless number of hydrogen \textgreater, num of radical electron is \textless number of radical electrons \textgreater, hybridization is \textless hybridization \textgreater, (is/is not) aromatic, (is/is not) in ring.}

Edge textual features: \textit{chemical bond. \textless bond type\textgreater bond, bond stereo is \textless bond stereo\textgreater, (is/is not) conjugated}

\section{Implementation Notes}
\label{appendix:imple}
\vpara{Running environment.}
All experiments are conducted on Linux machine with 945G RAM, and 8 NVIDIA A100 with 40GB GPU memory. For software versions, we use Python 3.11, Pytorch 2.0.1, DGL 1.1.2, transformers 4.32.1 and CUDA 11.8. Our code and datasets will be available.

\vpara{Hyper-parameters.}
The detailed pre-training hyper-parameters are listed in Table~\ref{tab:hyper}. 
For linear probing, we train the linear classifier using adam optimizer with lr=0.01 for 5000 epochs, and report the early-stopping results.
The detailed graph instruction tuning hyper-parameters are listed in Table~\ref{tab:git}.

\vpara{Baselines.}
To have a fair comparison, we download the public source code. For methods can not scale, we adapt their code to integrate with sampling algorithms to run on large-scale graphs. The sources of the codes used are as follows:
\begin{itemize}
    \item word2vec: \url{https://huggingface.co/fse/word2vec-google-news-300}
    \item DeBERTa-base: \url{https://huggingface.co/microsoft/deberta-base}
    \item DGI: \url{https://github.com/dmlc/dgl/blob/master/examples/pytorch/dgi/dgi.py}
    \item BRGL: \url{https://github.com/Namkyeong/BGRL\_Pytorch}
    \item GraphMAE: \url{https://github.com/THUDM/GraphMAE}
    \item GraphMAE2: \url{https://github.com/THUDM/GraphMAE2}
    \item GIANT-XRT: \url{https://github.com/amzn/pecos/tree/mainline/examples/giant-xrt}
    \item Prodigy: \url{https://github.com/snap-stanford/prodigy}
    \item OFA: \url{https://github.com/LechengKong/OneForAll}
    \item GraphGPT: \url{https://github.com/HKUDS/GraphGPT}
\end{itemize}

\vpara{Datasets splits.}
For Cora and PubMed, we follow commonly used data splits, using 20 labeled nodes per class as the training set, 30 nodes per class as the validation set, and the rest as the test set. We report the average accuracy on test set with 20 random initialization.

For Arxiv and Products, we follow the official splits~\cite{hu2020open}. Following the experimental procedure suggested by OGB, we repeat each experiment for 10 times with random seeds and report the average accuracy.

For Wiki-CS, we follow the official splits~\cite{mernyei2020wiki} with 20 different training splits, we report the average accuracy on the 20 different training splits with 20 random initialization. In each split, 5\% of the nodes in each class are used for training.

For FB15K237 and WN18RR, we follow splits in OFA~\cite{liu2021pre}. 
For FB15K237, training set has 272115 edges, validation set has 17535 edges and test set has 20466 edges.
For WN18RR, training set has 86835 edges, validation set has 3034 edges and test set has 3134 edges. We repeat each experiment for 10 times with random seeds and report the average accuracy.

For HIV and PCBA, we follow the official splits~\cite{hu2020open}.
We repeat each experiment for 10 times with random seeds and report the average accuracy.

\vpara{Linear probing.}
The dataset \(\mathcal{D}\) after generating embeddings, comprising embedding-label pairs \((\vh, y)\), is divided into training, validation, and test sets. 
A linear classifier with weight matrix \( \mW \in \mathbb{R}^{d \times |\mathcal{Y}|} \) is trained at top the embeddings from the frozen model, aiming to minimize the loss function \(\mathcal{L}\), typically cross-entropy, over the training set: \(\min_\mW \sum_{(\vh, y) \in \mathcal{D}_{\text{train}}} \mathcal{L}(\mW \cdot \vh, y)\). 
The performance of the model is evaluated based on a performance metric \( \mathcal{M} \), which can be defined generically as \(\mathcal{M}(\mathcal{D}_{\text{eval}}, f_{\theta}, \mW)\), where \(\mathcal{D}_{\text{eval}}\) refers to either the validation or test set. 

\vpara{Few-shot transfer.}
Our method follows our in-context learning approach in section~\ref{sec:icl}, and for baselines we either follow the same approach or use their already proposed in-context learning methods (Prodigy, OFA). We repeat each experiment for 10 times with random seeds and report the average accuracy.
All the other experimental details (pre-training) follow those for the previous experiment (i.e., linear probing). 

\vpara{Zero-shot transfer.}
For LLM-based baselines, we use the same prompts as our method without graph embeddings as input instruction prompts. The performance of zero-shot transfer is quantified using the accuracy of the LLM's generated text labels against the true labels.
For LLM base models like our method, Llama-7B and Llama2-7B, we take their outputs directly as predicted labels.
For LLM chat model like vicuna-7B-v1.5, we use regular expressions to extract predicted labels from its answers.


Graph Neural Networks (GNNs) are a class of deep learning models designed for processing data in graph form. 
A graph is defined by \( \mathcal{G} = (\mathcal{V}, \mathcal{E}) \), where \( \mathcal{V} \) denotes the set of nodes, each with a feature vector \( \vx_v \), and \( \mathcal{E} \) represents the edges, connecting two nodes, which may or may not have associated with a feature \( \vx_e \).

In GNNs, the feature vector of each node is iteratively updated based on the features of its neighboring nodes and the connecting edges.
The feature vector of a node \( v \) at the \( l \)-th layer, represented as \( \mH^{(l)}_v \), is updated as follows, initializing with \( \mH^{(0)}_v = \vx_v \):
\[
\mH^{(l+1)}_v = \sigma\left( f^{(l)}(\mH^{(l)}_v) + \sum_{u \in \mathcal{N}(v)} \phi\left( f^{(l)}(\mH^{(l)}_u), \vx_{e_{uv}} \right) \right),
\]
where \( \sigma \) denotes a non-linear activation function, \( f^{(l)}(*) \) represents a function applied at layer \( l \), \( \mathcal{N}(v) \) indicates the neighborhood of \( v \), and \( \phi \) combines the weighted features of neighboring nodes with edge features.
If edge features are not exist, \( \phi \) may solely rely on node features.
The combination function \( \phi \), incorporating edge features if available, is defined as:
\[
\phi\left( f^{(l)}(\mH^{(l)}_u), \vx_{e_{uv}} \right) = \alpha_{uv} \cdot \left( f^{(l)}(\mH^{(l)}_u) \odot \vx_{e_{uv}} \right),
\]
here, \( \alpha_{uv} \) is a scaling coefficient function, and \( \odot \) symbolizes an element-wise operation, such as multiplication or concatenation.

In our framework, when dealing with datasets containing edge text features (molecule graphs), we pre-process the edge text features using our language model.

\section{Extended Related Work}
\vpara{Single graph learning.}
Graph Neural Networks(GNNs)~\cite{kipf2016semi,velivckovic2018graph,yang2024testing} take node features as input and aggregate local neighbor representations using the message-passing paradigm, directly optimizing for specific downstream tasks, achieving superior performance.
With GNNs as the backbone, graph self-supervised learning~\cite{kipf2016variational,velivckovic2018deep} learns representation extractors on unlabeled graphs, subsequently applying the representations to downstream tasks or fine-tuning the pretrained GNNs. 
Unlike NLP tasks, graph learning tasks exhibit considerable diversity in their forms, making it a challenge to adapt a graph model to different downstream tasks. Recently, graph prompt learning has attempted to unify all tasks into either edge-level~\cite{liu2023graphprompt} or graph-level~\cite{sun2023all} tasks.

\vpara{Large language models on graphs.}
Many real-world graphs naturally come with text as node or edge features, which we refer to as Text-Attributed Graphs(TAGs). To facilitate the learning of graph models, using language models to encode text into low-dimensional vectors is a common practice~\cite{hu2020open}.
Furthermore, recent Large Language Models(LLMs), represented by ChatGPT, demonstrate extensive common knowledge and capabilities as general task solvers. 
Some works transform graph structures into text and combine them with task descriptions to form prompts as input for LLMs, attempting to directly use LLMs as predictors for handling graph learning tasks~\cite{guo2023gpt4graph, wang2023can, liu2023evaluating,sui2024knowledgegraphsmakelarge}. 
Furthermore, some research attempts to enhance the ability of LLMs to understand graph structures and tasks by employing instruction tuning~\cite{ye2023natural, tang2023graphgpt,xiong2024large,yang2023harnessing,sui2024fidelis}.

\section{Instruction Prompts}
\label{appendix:prompts}
\vpara{Citation networks.}
\textit{Given a citation graph, node represents academic paper with a specific topic. \textless $\text{node}_v$\textgreater \space is featured with its content: [Title], [Abstract]. \textless $\text{node}_v$\textgreater \space and its contextual neighbor nodes \{$\textless \text{node}_u\textgreater; u \in \mathcal{V}$\} are highly correlated. Question: Which category should \textless $\text{node}_v$\textgreater \space be classified as? Please strictly classify the paper into one of the following categories:[Candidate Labels].  Answer: }

\vpara{Products networks.}
\textit{Given a products graph, node represents a product sold in Amazon with a specific category. \textless $\text{node}_v$\textgreater \space is featured with its content: [Content]. \textless $\text{node}_v$\textgreater \space and its contextual neighbor nodes \{$\textless \text{node}_u\textgreater; u \in \mathcal{V}$\} are highly correlated. Question: Which category should \textless $\text{node}_v$\textgreater \space be classified as? Please strictly classify the product into one of the following categories:[Candidate Labels].  Answer: }

\vpara{Web networks.}
\textit{Given a Wikipedia graph, node represents Wikipedia page with a specific category. \textless $\text{node}_v$\textgreater \space is featured with its content: [Name],[Content]. \textless $\text{node}_v$\textgreater \space and its contextual neighbor nodes \{$\textless \text{node}_u\textgreater; u \in \mathcal{V}$\} are highly correlated. Question: Which category should \textless $\text{node}_v$\textgreater \space be classified as? Please strictly classify the Wikipedia page into one of the following categories:[Candidate Labels].  Answer: }

\vpara{Knowledge graphs.}
\textit{Given a knowledge graph, edge between two entities represents a relation with a specific category. Node one \textless $\text{node}_v$\textgreater \space is featured with its content: [Name],[Content]. Node two \textless $\text{node}_u$\textgreater \space is featured with its content: [Name],[Content]. Question: Which category should the relation between node one Node one \textless $\text{node}_v$\textgreater \space and node two \textless $\text{node}_u$\textgreater \space be classified as? Please strictly classify the relation into one of the following categories:[Candidate Labels].  Answer: }

\end{document}